\documentclass[runningheads]{llncs}

\usepackage{graphicx}
\usepackage{comment}
\usepackage{amsmath,amssymb}
\usepackage{color}
\usepackage{url}
\usepackage{hyperref}

\newif\ifreview
\reviewfalse

\ifreview
	\usepackage{lineno}

	\linenumbers
\fi

\usepackage[table,xcdraw,dvipsnames]{xcolor}
\usepackage[acronym, shortcuts]{glossaries}

\newacronym{ml}{ML}{Machine Learning}
\newacronym{ai}{AI}{Artificial Intelligence}
\newacronym{dl}{DL}{Deep Learning}
\newacronym{xai}{XAI}{eXplainable AI}

\newacronym{eo}{EO}{Earth Observation}
\newacronym{rs}{RS}{Remote Sensing}
\newacronym{scl}{SCL}{Scene Classification Layer}
\newacronym{ads}{ADS}{additional data sources}
\newacronym{rgb}{RGB}{Red-Green-Blue}
\newacronym{dem}{DEM}{Digital Elevation Maps}
\newacronym{srtm}{SRTM}{Shuttle Radar Topography Mission}
\newacronym{aster}{ASTER}{Advanced Spaceborne Thermal Emission and Reflection Radiometer}
\newacronym{alos}{ALOS}{Advanced Land Observing Satellite}
\newacronym{ecmwf}{ECMWF}{European Center for Medium-Range Weather Forecasts}
\newacronym{sar}{SAR}{Synthetic Aperture Radar}
\newacronym{lidar}{LiDAR}{Light Detection and Ranging}
\newacronym{fapar}{FAPAR}{Fraction of Absorbed Photosynthetically Active Radiation}
\newacronym{twi}{TWI}{Topographic Wetness Index}
\newacronym{s2}{S2}{Sentinel-2}
\newacronym{sits}{SITS}{satellite image time series}
\newacronym{gsd}{GSD}{ground sampling distance}
\newacronym{lulc}{LULC}{Land Use and Land Cover}

\newacronym{ndvi}{NDVI}{normalized difference vegetation index}
\newacronym{evi}{EVI}{enhanced vegetation index}

\newacronym{rf}{RF}{random forest}
\newacronym{gpr}{GPR}{gaussian process regression}
\newacronym{svm}{SVM}{support vector machine}
\newacronym{mlp}{MLP}{multilayer perceptron}
\newacronym{dnn}{DNN}{deep neural network}
\newacronym{rnn}{RNN}{Recurrent neural network}
\newacronym{cnn}{CNN}{convolutional neural network}
\newacronym{1d-cnn}{1D-CNN}{1-Dimensional convolutional neural network}
\newacronym{dfnn}{DFNN}{deep forward neural network}
\newacronym{lstm}{LSTM}{long short-term memory}
\newacronym{alstm}{ALSTM}{attention-based long short-term memory}
\newacronym{gbdt}{GBDT}{gradient-boosted decision tree}
\newacronym{pca}{PCA}{Principal Component Analysis}
\newacronym{mha}{MHA}{multi-head self-attention}
\newacronym{sdpa}{SDPA}{scaled dot-product attention}

\newacronym{mae}{MAE}{mean absolute error}
\newacronym{mse}{MSE}{mean squared error}
\newacronym{rmse}{RMSE}{root mean square error}
\newacronym{rrmse}{RRMSE}{relative root mean square error}
\newacronym{r2}{$\text{R}^2$}{coefficient of determination}
\newacronym{pp}{p.p}{percentage points}
\newacronym{iou}{IoU}{intersection over union}

\setacronymstyle{long-short}
\usepackage{marvosym}
\usepackage{stix}
\usepackage{multirow} 
\usepackage{adjustbox}
\usepackage[normalem]{ulem}
\useunder{\uline}{\ul}{}

\newcommand{\benge}{Benge }
\newcommand{\treesat}{TreeSAT }
\newcommand{\yc}{CropYield }

\newcommand\inmod{\textcolor{ForestGreen}{\MVRightArrow $\laplac$}}
\newcommand\outmod{\textcolor{RedOrange}{$\laplac \text{\MVRightArrow}$}}

\begin{document}

%%%%%%%%%%%%%%%%%%%%% Add submission id, track, and title. %%%%%%%%%%%%%%%%%%%%%

% TODO: Please insert your submission number here
\def\SubNumber{072}

% TODO: Please uncomment the track this paper will be submitted to, comment all other lines
%\def\GCPRTrack{Main Track}
%\def\GCPRTrack{Special Track: Pattern recognition in the life and natural sciences}
\def\GCPRTrack{Special Track: Photogrammetry and remote sensing}
%\def\GCPRTrack{Young Researcher's Forum}
%\def\GCPRTrack{Fast Review Track}

% TODO: Replace with your title
\title{Can Multitask Learning Enhance Model Explainability?\thanks{H.Najjar acknowledges support through a scholarship from the University of Kaiserslautern-Landau.
    The research results presented are part of a large collaborative project on agricultural yield predictions, which was partly funded through the ESA InCubed Programme (\url{https://incubed.esa.int/}) as part of the project AI4EO Solution Factory (\url{https://www.ai4eo-solution-factory.de/}).}}
% You can use \thanks for acknowledgment. Do not add any acknowledgment to the draft 
% version that is used for the review process.  
%\title{Title\thanks{XXX}}

\ifreview
	% ANONYMOUS SUBMISSION FOR REVIEW
	% DO NOT MODIFY these for the draft version that is used for the review process.
	\titlerunning{GCPR 2025 Submission \SubNumber{}. CONFIDENTIAL REVIEW COPY.}
	\authorrunning{GCPR 2025 Submission \SubNumber{}. CONFIDENTIAL REVIEW COPY.}
	\author{GCPR 2025 - \GCPRTrack{}}
	\institute{Paper ID \SubNumber}
\else
	% If the paper title is too long for the running head, you can set an abbreviated paper title here

	\author{Hiba Najjar\inst{1,2}\orcidID{0000-0002-7498-794X} \and
	Bushra Alshbib\inst{1} \and
	Andreas Dengel\inst{1,2}\orcidID{0000-0002-6100-8255}}
	
	\authorrunning{H. Najjar et al.}
	
	\institute{Kaiserslautern-Landau University, Kaiserslautern, Germany \\ \email{alshbib@rptu.de} \and German Research Center for Artificial Intelligence, Kaiserslautern, Germany
	\email{\{hiba.najjar,andreas.dengel\}@dfki.de}}
\fi

\maketitle % typeset the header of the contribution

\begin{abstract}
%The abstract should briefly summarize the contents of the paper in 150--250 words.
Remote sensing provides satellite data in diverse types and formats. The usage of multimodal learning networks exploits this diversity to improve model performance, except that the complexity of such networks comes at the expense of their interpretability.
In this study, we explore how modalities can be leveraged through multitask learning to intrinsically explain model behavior. 
In particular, instead of additional inputs, we use certain modalities as additional targets to be predicted along with the main task. The success of this approach relies on the rich information content of satellite data, which remains as input modalities. We show how this modeling context provides numerous benefits:    
    (1) in case of data scarcity, the additional modalities do not need to be collected for model inference at deployment, 
    (2) the model performance remains comparable to the multimodal baseline performance, and in some cases achieves better scores,
    (3) prediction errors in the main task can be explained via the model behavior in the auxiliary task(s).
We demonstrate the efficiency of our approach on three datasets, including segmentation, classification, and regression tasks.
Code available as supplementary material and at \url{git.opendfki.de/hiba.najjar/mtl_explainability/}.

\end{abstract}

\section{Introduction}\label{sec:intro}

    Multimodal learning is widely used across various fields, driven by the availability of data from diverse sources or sensors. \gls{rs} benefits particularly from this field, as it provides a wide range of satellite images and satellite-derived products.
    In fact, it was shown that models fusing data from different modalities outperform their uni-modal counterparts both intuitively and provably \cite{huang2021makes}.
    To adjust to the multimodal setup, advanced modeling techniques are often implemented. However, these techniques often lead to increased model complexity, which comes at the expense of model interpretability \cite{joshi2021review,gunther2024explainable}.

    In contrast, multitask learning aims at predicting multiple targets using a shared model, achieving in most cases smaller memory footprint, reduced number of calculations, and improved performance \cite{maniscalco2024multimodal,ding2019effectiveness,lu202012,kendall2018multi,sener2018multi,liu2019end,levering2021relation,zhang2021survey,vandenhende2021multi}. 
    There are still certain scenarios in which single task networks might outperform multitask counterparts, due to the number of tasks, their types, and the accuracy of their annotated labels \cite{vandenhende2021multi,standley2020tasks,narazani2022pet,thomason2019shifting}.
        
    In this study, we investigate a specific approach to explaining model predictions in the context of multimodal learning by using the framework of multitask learning. By treating certain modalities as auxiliary tasks, we achieve two key objectives: first, we mitigate the model's dependence on these modalities during training, as such modalities are no longer required as inputs during deployment. Second, we provide insights into model behavior by analyzing prediction errors and accuracies across multiple tasks, in order to intrinsically interpret multimodal networks. %When these errors exhibit correlation, they offer a means to explain model behavior.

\section{Related work}\label{sec:background}

    \subsection{Explainable multimodal networks}
    
        \Gls{xai} research line provides various techniques to tackle the interpretability of neural networks and achieves various objectives. A common goal of \gls{xai} is \textit{Justification} \cite{adadi2018peeking}, answering the question \textit{"Why did the model make this prediction?"}. Feature attribution methods, for instance, measure the influence of each single (or group of) input feature(s) on the prediction \cite{lime_ribeiro2016should,shap_lundberg2017unified,ig_sundararajan2017axiomatic,selvaraju2017grad}. Many such methods are model-agnostic, and can thus be readily applied to multimodal networks, but they are likely less accurate than intrinsic methods, which rely on the model's internal elements to explain its behavior.
        Another goal of \gls{xai}, less commonly addressed, is \textit{Control}, consisting of understanding model errors, ultimately leading to improving its model reasoning and avoiding more errors \cite{adadi2018peeking}.
        In this manuscript, we aim at achieving this goal through an intrinsic technique which leverages multitask learning.
        
    \subsection{Explainability through multitask learning}
        
        % joint training
            Among the few intrinsic methods in \gls{xai} based on multitask learning is \textbf{joint training}, which generates explanations by augmenting the original network with additional tasks to explicitely return textual, imagery or numerical explanations, along with the model's main decision \cite{hendricks2016generating,park2018multimodal,liu2019towards,kanehira2019learning,rio2020understanding,liu2020deep,tang2023takes}.
            Park et al. \cite{park2018multimodal} introduce a framework for image classification tasks which generates textual and visual explanations as auxiliary tasks. Their main limitation is the necessity of an annotated explanation dataset, which should include text explanations and attention maps.
            Hendricks et al. \cite{hendricks2016generating} apply joint training to explain a classification task of bird species, by predicting the class label and a corresponding textual explanation. 
            Although their proposed method relies on reinforcement learning, it requires textual annotations of the training dataset.
            Similarly, Rio et al. \cite{rio2020understanding} also proposes a network which returns visual explanation to the classification task, yet relies as well on bounding boxes around the object to be classified, to learn the explanations in a weakly-supervised manner.
            
        % semantic bottlenecks
            Another line of explanation methods close to the joint training family are \textbf{semantic bottleneck networks}. Such models were introduced by Losch et al.  \cite{losch2019interpretability} and consist of defining an intermediate \textit{bottleneck} layer where latent features are enforced to align with semantic concepts. 
            %The proposed encoder-decoder design has the downside of non-linearity between the representations learned at the bottleneck layer and the model final predictions. To address this issue and improve model interpretability, another study proposed to place the semantic layer right before the final layer, enabling a linear mapping between the concepts and the predictions \cite{marcos2019semantically}. A similar approach was applied in different fields.
            A study improved this method and proposed to place the semantic layer right before the final layer, enabling a linear mapping between the concepts and the predictions \cite{marcos2019semantically}. 
            This approach was applied in different applications in remote sensing \cite{levering2021relation}, healthcare \cite{mojab2019deep}, and autonomous driving \cite{echterhoff2024driving}.
          
        % our approach
            While previous studies propose techniques to explicitly predict explanations for model predictions, they are often limited by the availability of annotations for the explanation task, in the form of semantic labels for the semantic bottleneck approach, or explicit sentences and scores for the joint training framework. In our work, we overcome this limitation by relying on available input modalities and turn them into explanatory auxiliary tasks. While this method does not provide explicit explanations, we explore how to extract insightful results from this framework to intrinsically explain model predictions and errors for three different tasks.

\section{Methodology}\label{sec:methods}

    \subsection{Interpretability through Multitask Learning}

        \begin{figure}[t]
            \centering
            \adjustbox{center=\textwidth}{\includegraphics[width=0.9\textwidth]{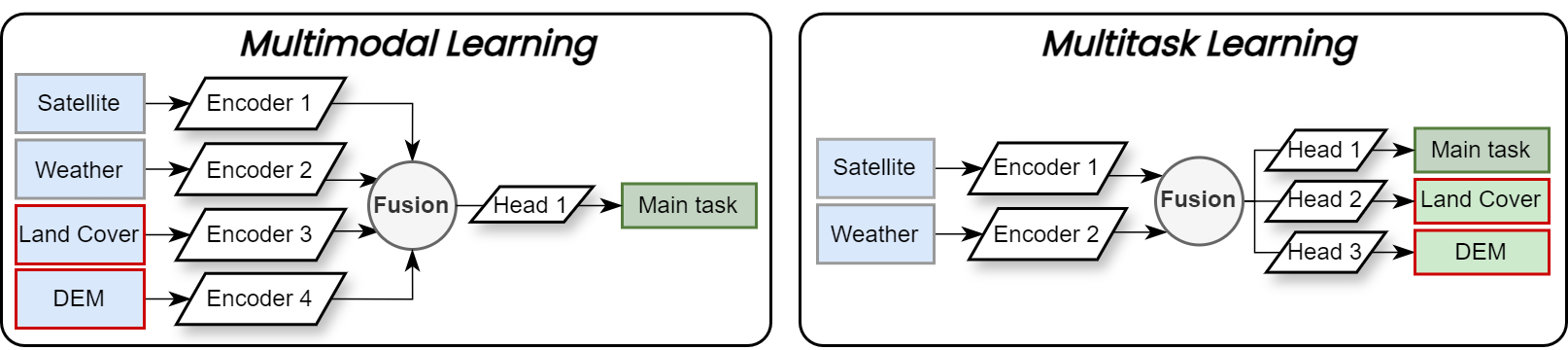}}
            \caption{Comparison of multimodal against multitask setups in a \gls{rs} dataset. DEM refers to digital elevation maps.}
            \label{fig:mml_mtl}
        \end{figure}

        %We propose leveraging multimodal datasets to enhance model interpretability via a multitask learning framework. 
        Additional modalities in multimodal datasets are typically incorporated as input data, yet not all of them may be essential for achieving the baseline model performance. In particular, satellite imagery inherently encodes a rich and diverse range of information about the Earth's surface. For instance, multispectral sensors capture spectral characteristics across multiple bands, while \gls{sar} sensors provide structural and textural details. 
        Exploiting this characteristic of satellite data, we focus on \gls{rs} multimodal datasets and explore the effect of shifting auxiliary modalities between input data and auxiliary tasks, as depicted in Figure~\ref{fig:mml_mtl}, analyzing its impact on both model performance and interpretability. 
        To maintain a robust baseline, we ensure that satellite data remains an input modality in all multitask experiments, to avoid significant performance degradation. We evaluate our approach on the following three datasets.

    \subsection{Datasets}

        \subsubsection{\yc for yield prediction}
            The \textit{\yc} dataset contains approximately 500 crop yield maps of corn, soybean, and wheat fields located in Northern Argentina, covering crop seasons from 2017 to 2023.
            Since the dataset is processed on a pixel-wise basis, it contains more than 3.5 million input samples. 
            The available modalities include satellite multispectral imagery, weather data, \gls{dem} properties, and crop type. Both satellite and weather data are temporal, spanning from seeding to harvesting dates each year. Yield maps, rasterized at a 10-meter resolution, are used as the main target (regression task).
            Further details are provided in Appendix~\ref{app:data}.
            Due to confidentiality restrictions, this dataset cannot be publicly released.

        \subsubsection{\benge for land cover segmentation}
            % https://github.com/HSG-AIML/ben-ge
            \textit{\benge} is an open-source multimodal dataset for \gls{lulc} segmentation, extending the BigEarthNet dataset \cite{mommert2023ben,sumbul2019bigearthnet,sumbul2021bigearthnet}. 
            It contains \gls{sar} and multispectral satellite images, from Sentinel-1 and Sentinel-2 missions respectively, for 590,326 locations throughout Europe, complemented with elevation maps, environmental data, climate zone information, and seasonal encoding.
            Following the recommendations of \cite{mommert2023ben}, our experiments were initially conducted on a small subset of the dataset. Subsequently, the best-performing architectures were trained on the 0.2 split of the full dataset, in order to balance computational efficiency with comparable performance.

        \subsubsection{\treesat for tree identification}
            \textit{\treesat} is an open-source dataset for tree species classification in Central Europe based on multi-sensor data from aerial imagery and satellite observations, including \gls{sar} and multispectral images \cite{ahlswede2022treesatai}. The dataset contains labels of 15 tree genera (the main classification task), nine forest stand types, and three foliage types, corresponding to classification levels L3, L2, and L1, respectively.
            Additionally, it includes an approximation of tree age, which is treated as a continuous feature. 

    \subsection{Experimental Setup}

        \subsubsection{Modality Encoders}
            %We evaluate and compare the performance of both multimodal and multitask learning on multimodal datasets. 
            Given the diversity of the input data types, we adopt an intermediate fusion approach: each input modality is processed by a dedicated encoder, generating an intermediate representation, which is then fused across modalities before being passed to a task-specific head for the final predictions.
            This approach facilitates handling multiple input modalities despite differences in data type, spatial characteristics, and temporal resolutions.
            It has also often outperformed early and late fusion techniques in \gls{rs} applications \cite{mena2024common}.
            The architecture of the encoder is chosen based on the types of the input and the target:
            For \text{imagery inputs}, we either use a U-Net architecture in segmentation tasks or a convolutional network in other tasks. If the input image is small, such as in low-resolution satellite imagery, we flatten it and process it using a \gls{mlp}. 
            \text{Time-series inputs} are processed using Transformers, including positional encoding based on each timestamp.
            \text{Tabular data} are processed using \glspl{mlp}, whether they include a single or multiple features.
            Finally, for \text{categorical inputs}, we use an \gls{mlp} or an embedding layer.

        \subsubsection{Fusion Block}
            The intermediate representations generated by the modality encoders are combined at the fusion block through concatenation, optionally followed by convolutional layers:
            For \text{regression and classification tasks}, each encoder outputs a one-dimensional feature vector representing its respective modality. These vectors are simply concatenated at the fusion stage, with no additional processing.
            For \text{segmentation tasks}, modalities are encoded into a three-dimensional latent representation (i.e., channels $\times$ height $\times$ width). If the input is an image processed via a U-Net, this representation is obtained naturally. For tabular data encoded through a \gls{mlp}, the one-dimensional output can be expanded into additional dimensions to align with the spatial structure of other representations. This alignment facilitates the concatenation along the channel dimension, followed by additional convolutional layers that preserve the spatial characteristics (height and width) of the fused representation.

        \subsubsection{Prediction Heads}
            Multiple prediction heads can branch out from the fusion block, each dedicated to a specific target:
            For \text{segmentation tasks}, the prediction head consists of convolutional layers, which preserve the spatial dimensions of the image.
            For \text{regression and classification tasks}, a \gls{mlp} is used to return the appropriate number of output neurons for the task.

        \subsubsection{Loss and Metrics}
            The optimization loss for each task is defined based on its nature. For \text{classification tasks}, including semantic segmentation, the cross-entropy loss is used, whereas for \text{regression tasks}, including dense segmentation, we use the \gls{mse} function. 
            In the multitask learning scenario, the loss contributions of individual tasks are manually fine-tuned. For example, we evaluated strategies such as equally distributing the loss contribution across all tasks, or prioritizing the primary task by assigning it a higher weight (e.g., 60\% or 80\%) while maintaining a uniform distribution of weights across auxiliary tasks.
            To further evaluate and report performance, additional metrics are included. \Gls{mae} and \gls{r2} are used for regression and dense segmentation tasks, the F1 score for classification tasks, and the \gls{iou} for semantic segmentation tasks.\\

        In Table~\ref{tab:modalities} in Appendix~\ref{app:data}, we provide a summary of the encoder, prediction head, loss function, and evaluation metric used for each modality in each dataset.

\section{Results}\label{sec:results}

    \subsection{Multimodal vs. Multitask modeling}

        In this section, we analyze the performance results of the different modeling setups, including baselines, which include the remotely sensed images (aerial and satellites) and temporal modalities, multimodal learning experiments (MML), which test different combinations of additional input modalities, and multitask learning experiments (MTL), which shifts some modalities from being additional input to auxiliary targets. \\

        Starting with the \yc dataset, Table~\ref{tab:yc_res} combines the results of the main experiments. Table~\ref{tab:yc_res_full} in Appendix~\ref{app:yc_mod} contains more results, particularly extending the baseline experiments.
            In multimodal setups, performance comparable to the baseline is observed when including weather and \gls{dem} as additional inputs to the model, in Experiment \textbf{5}, while any other combination of auxiliary inputs yields a decline in the performance. Surprisingly, this includes Experiments \textbf{2}, \textbf{4}, and \textbf{6}, where we provide the model with the crop label of each pixel sample. In contrast, forcing the model to predict this label improved its performance, particularly when including weather and \gls{dem} modalities as inputs, in Experiment \textbf{9}, and when including no additional input modality, in Experiment \textbf{7}. The latter even reached the highest overall \gls{r2} score across all experiments.
            The model further reached a very high F1-score of 99.4\%  in the crop classification task, which brings a great benefit in practice, enabling the distinction of crop types along the accurate yield prediction.
            We assume that the performance gap in yield prediction between Experiments \textbf{2} and \textbf{7} is due to the shared representation of the multitask learning setup, in which the model is forced to learn representations related to the different crop labels, which positively influences the accuracy of the predicted yield. 
            In the explainability analysis, we will focus on Experiment \textbf{7}, which predicts the yield and crop labels using the satellite data alone.\\

        \begin{table}[t]
        \caption{Modeling performance on the test set of the \yc dataset. The best and second-best scores are highlighted in bold and underlined, respectively. Crop classification performance is given in micro F1 score.}
        \centering
        \adjustbox{center=\textwidth}{
        \begin{tabular}{cccccccccccccc}
            \hline
             & \multicolumn{2}{c}{} &  & \multicolumn{4}{c}{\textbf{Modalities}} &  & \textbf{Main task} &  & \multicolumn{2}{c}{\textbf{Auxiliary tasks}} &  \\ \cline{5-8} \cline{10-10} \cline{12-13}
            \quad \textbf{ } & \multicolumn{2}{c}{\textbf{Experiments}} & \textbf{ } & \textbf{Satellite } & \textbf{ Crop label } & \textbf{ Weather } & \textbf{ DEM} &  \textbf{ } & \textbf{\begin{tabular}[c]{@{}c@{}}Yield\\ (R²)\end{tabular}} & \textbf{ } & \textbf{\begin{tabular}[c]{@{}c@{}}Crop cls.\\ (F1)\end{tabular}} & \textbf{\begin{tabular}[c]{@{}c@{}}DEM\\ (MAE)\end{tabular}} & \quad \textbf{ } \\ \cline{2-3} \cline{5-8} \cline{10-10} \cline{12-13}
             & Baseline & 1 &  & \inmod &  &  &  &  & {\ul 0.81} &  & - & - &  \\ \cline{2-3} \cline{5-8} \cline{10-10} \cline{12-13}
             & MML & 2 &  & \inmod & \inmod &  &  &  & 0.77 &  & - & - &  \\
             &  & 3 &  & \inmod &  & \inmod &  &  & 0.75 &  & - & - &  \\
             &  & 4 &  & \inmod & \inmod & \inmod &  &  & 0.79 &  & - & - &  \\
             &  & 5 &  & \inmod &  & \inmod & \inmod &  & {\ul 0.81} &  & - & - &  \\
             &  & 6 &  & \inmod & \inmod & \inmod & \inmod &  & 0.79 &  & - & - &  \\ \cline{2-3} \cline{5-8} \cline{10-10} \cline{12-13}
             & MTL & 7 &  & \inmod & \outmod &  &  &  & \textbf{0.82} &  & {\ul 99.4} & - &  \\
             &  & 8 &  & \inmod & \outmod & \inmod &  &  & 0.77 &  & \textbf{99.5} & - &  \\
             &  & 9 &  & \inmod & \outmod & \inmod & \inmod &  & 0.80 &  & \textbf{99.5} & - &  \\
             &  & 10 &  & \inmod & \outmod & \inmod & \outmod &  & 0.75 &  & 99.3 & \textbf{0.42} &  \\ \hline
        \end{tabular}
        }
        \adjustbox{center=\textwidth}{\parbox{1.5\textwidth}{\centering \small   \inmod \textbf{} Input \quad | \quad \outmod \textbf{} Output \quad | \quad cls.:classification.}}
        \label{tab:yc_res}
        \end{table}

        Moving to the \benge dataset, we present the results in Table~\ref{tab:benge_res}.
        The complete table including model performance on auxiliary tasks is presented in Table~\ref{tab:benge_res_full} in Appendix~\ref{app:benge_mod}.
            In the baseline experiment, the model is trained on the multispectral and \gls{sar} satellite images alone, achieving the second best scores in the main task of \gls{lulc}, with an accuracy of 87.94\% and an \gls{iou} score of 0.388.
            In the multimodal Experiments (\textbf{2-8}), we evaluate different combinations of one or more additional input modalities, prioritizing elevation data due to its spatial dimension, which the remaining modalities lack. While all multimodal experiments yielded results comparable to the baseline, Experiment \textbf{7} including the elevation and weather data have slighlty outperformed it, achieving an accuracy of 87.95\%. Similarly, Experiment \textbf{4}, which includes seasonal information,  achieves a marginally higher \gls{iou} score of 0.389, also surpassing the baseline.
            In the multitask setup, the \gls{lulc} accuracies remain within a similar range, while \gls{iou} scores marginally declined. Notably, certain modality combinations reached improved accuracies when incorporated as auxiliary tasks rather than as input modalities, such as climate zone (in Experiments \textbf{3} and \textbf{10}) and the combination of all modalities (in Experiments \textbf{8} and \textbf{15}).
            Overall, we find that the additional input modalities do not contribute to improved model performance. However, our results remain consistent with the scores reported in \cite{mommert2023ben}. Moreover, the multitask setup neither degrades nor enhances the primary task’s performance, while its other benefits persist.
            In Section~\ref{ssec:xai}, we further investigate the explanatory capacity of each output modality, using Experiment \textbf{15} as a testbed.\\

        \begin{table}[t]
        \caption{Test set performance on the \benge dataset. The best and second-best scores are highlighted in bold and underlined, respectively. Climate zone classification performance is given in micro F1 score.}
        \adjustbox{center=\textwidth}{
        \begin{tabular}{ccccccccccccc}
        \hline
         & \multicolumn{2}{c}{} &  & \multicolumn{5}{c}{\textbf{Modalities}} &  & \multicolumn{2}{c}{\textbf{Main task}} &  \\ \cline{5-9} \cline{11-12}
        \quad \textbf{ } & \multicolumn{2}{c}{\textbf{Experiment}} & \quad \textbf{ } & \textbf{Satellite} & \textbf{Elevation} & \textbf{\begin{tabular}[c]{@{}c@{}}Climate\\ Zone\end{tabular}} & \textbf{Season} & \textbf{Weather} & \quad \textbf{ } & \textbf{\begin{tabular}[c]{@{}c@{}}LULC\\ (Accuracy)\end{tabular}} & \textbf{\begin{tabular}[c]{@{}c@{}}LULC\\ (IoU)\end{tabular}} & \quad \textbf{ } \\ \cline{2-3} \cline{5-9} \cline{11-12}
         & Baseline & 1 &  & \inmod &  &  &  &  &  & {\ul 87.94} & {\ul 0.388} &  \\ \cline{2-3} \cline{5-9} \cline{11-12}
         & MML & 2 &  & \inmod & \inmod &  &  &  &  & 87.91 & 0.386 &  \\
         &  & 3 &  & \inmod &  & \inmod &  &  &  & 87.90 & 0.386 &  \\
         &  & 4 &  & \inmod &  &  & \inmod &  &  & 87.91 & \textbf{0.389} &  \\
         &  & 5 &  & \inmod &  &  &  & \inmod &  & 87.93 & 0.387 &  \\
         &  & 6 &  & \inmod & \inmod &  & \inmod &  &  & 87.90 & 0.385 &  \\
         &  & 7 &  & \inmod & \inmod &  &  & \inmod &  & \textbf{87.95} & 0.383 &  \\
         &  & 8 &  & \inmod & \inmod & \inmod & \inmod & \inmod &  & 87.85 & 0.387 &  \\ \cline{2-3} \cline{5-9} \cline{11-12}
         & MTL & 9 &  & \inmod & \outmod &  &  &  &  & 87.90 & 0.380 &  \\
         &  & 10 &  & \inmod &  & \outmod &  &  &  & 87.93 & 0.379 &  \\
         &  & 11 &  & \inmod &  &  & \outmod &  &  & 87.91 & 0.380 &  \\
         &  & 12 &  & \inmod &  &  &  & \outmod &  & 87.91 & 0.381 &  \\
         &  & 13 &  & \inmod & \outmod &  & \outmod &  &  & 87.91 & 0.377 &  \\
         &  & 14 &  & \inmod & \outmod &  &  & \outmod &  & 87.89 & 0.377 &  \\
         &  & 15 &  & \inmod & \outmod & \outmod & \outmod & \outmod &  & 87.89 & 0.373 &  \\ \hline
        \end{tabular}
        }
        \label{tab:benge_res}
        \end{table}

            \treesat dataset exhibits different patterns, as shown in the results displayed in Table~\ref{tab:treesat_res}.
            The baseline model, trained on the three imagery modalities (i.e. aerial imagery and two satellite images), achieves a micro F1-score of 74.3\%, ranking second. Using the same best-performing model architecture, this represents a significant improvement compared to the 71.66\% accuracy reported by Ahlswede et al. \cite{ahlswede2022treesatai}.
            As shown in Table~\ref{tab:treesat_res}, the highest accuracy of 76.9\% is reached by the multimodal experiment that includes the age as an additional input data.  Tree type labels from levels 1 and 2 were not included as input features, as acquiring this data at inference time would be impractical in real-world scenarios.
            In contrast, age can, in some cases, be inferred from historical records and old maps which document events such as deforestation, wildfires, or planting.
            In the multitask experiments, the primary task's performance declines slightly but maintains F1-scores above 70\%. Specifically, Experiment \textbf{4}, which predicts only the first level (L1), and Experiment \textbf{8}, which infers all modalities, yield the lowest L1 F1-scores of 70.3\% and 70.4\%, respectively. In contrast, including the second level (L2) in Experiment \textbf{3} achieved the same accuracy as the baseline model (74.3\%) while also yielding accurate labels for the second level labels, reaching a micro F1-score of 78.2\%.
            Overall, in the multitask experiments, L2 classification (with 9 classes) demonstrates high accuracy, L1 classification (with 3 classes) achieves significantly better scores, while age prediction (with normalized values) exhibits moderate performance.
            Experiment \textbf{7}, which reached the second performance in the main task among multitask experiments, will be explored in the explanatory analysis in the following section.
                % age: normalized

        \begin{table}[t]
        \caption{Test set performance on the TreeSAT dataset. The best and second-best scores are highlighted in bold and underlined, respectively. \textit{Images} refer to the aerial and two satellite images (from Sentinel-1 and Sentinel-2 missions). L3, L2, and L1 classification performance are given in micro F1 score.}
        \adjustbox{center=\textwidth}{
        \begin{tabular}{cclccccccccccccccc}
        \cline{1-17}
         & \textbf{} &  &  & \multicolumn{4}{c}{\textbf{Modalities}} &  & \textbf{Main task} &  & \multicolumn{5}{c}{\textbf{Auxiliary tasks}} &  &  \\ \cline{5-8} \cline{10-10} \cline{12-16}
        \quad \textbf{ } & \multicolumn{2}{c}{\textbf{Experiment}} & \textbf{ } & \textbf{Images} & \textbf{L2} & \textbf{L1} & \textbf{Age} &  \textbf{ } & \textbf{L3 (F1)} &  \textbf{ } & \textbf{L2  (F1)} &  & \textbf{L1 (F1)} &  & \textbf{Age (MAE)} & \quad \textbf{ } & \textbf{} \\ \cline{2-3} \cline{5-8} \cline{10-10} \cline{12-16}
         & Baseline & 1 &  & \inmod &  &  &  &  & {\ul 74.3} &  &  &  &  &  &  &  & {\ul } \\ \cline{2-3} \cline{5-8} \cline{10-10} \cline{12-16}
         & MML & 2 &  & \inmod &  &  & \inmod &  & \textbf{76.9} &  &  &  &  &  &  &  & \textbf{} \\ \cline{2-3} \cline{5-8} \cline{10-10} \cline{12-16}
         & MTL & 3 &  & \inmod & \outmod &  &  &  & {\ul 74.3} &  & \textbf{78.2} &  &  &  &  &  & \textbf{} \\
         &  & 4 &  & \inmod &  & \outmod &  &  & 70.3 &  &  &  & {\ul 92.1} &  &  &  & {\ul } \\
         &  & 5 &  & \inmod &  &  & \outmod &  & 71.8 &  &  &  &  &  & \textbf{0.52} &  &  \\
         &  & 6 &  & \inmod & \outmod & \outmod &  &  & 71.1 &  & 76.6 &  & \textbf{92.3} &  &  &  & \textbf{} \\
         &  & 7 &  & \inmod & \outmod &  & \outmod &  & 72.2 &  & {\ul 77.3} &  &  &  & \textbf{0.52} &  & {\ul } \\
         &  & 8 &  & \inmod & \outmod & \outmod & \outmod &  & 70.4 &  & 75.5 &  & {\ul 92.2} &  & {\ul 0.53} &  &  \\ \cline{1-17}
        \end{tabular}
        }
        \label{tab:treesat_res}
        \end{table}

    \subsection{Model Explainability}\label{ssec:xai}

        \subsubsection{\yc}

\begin{figure}[t]
            \centering
            \includegraphics[width=0.85\textwidth]{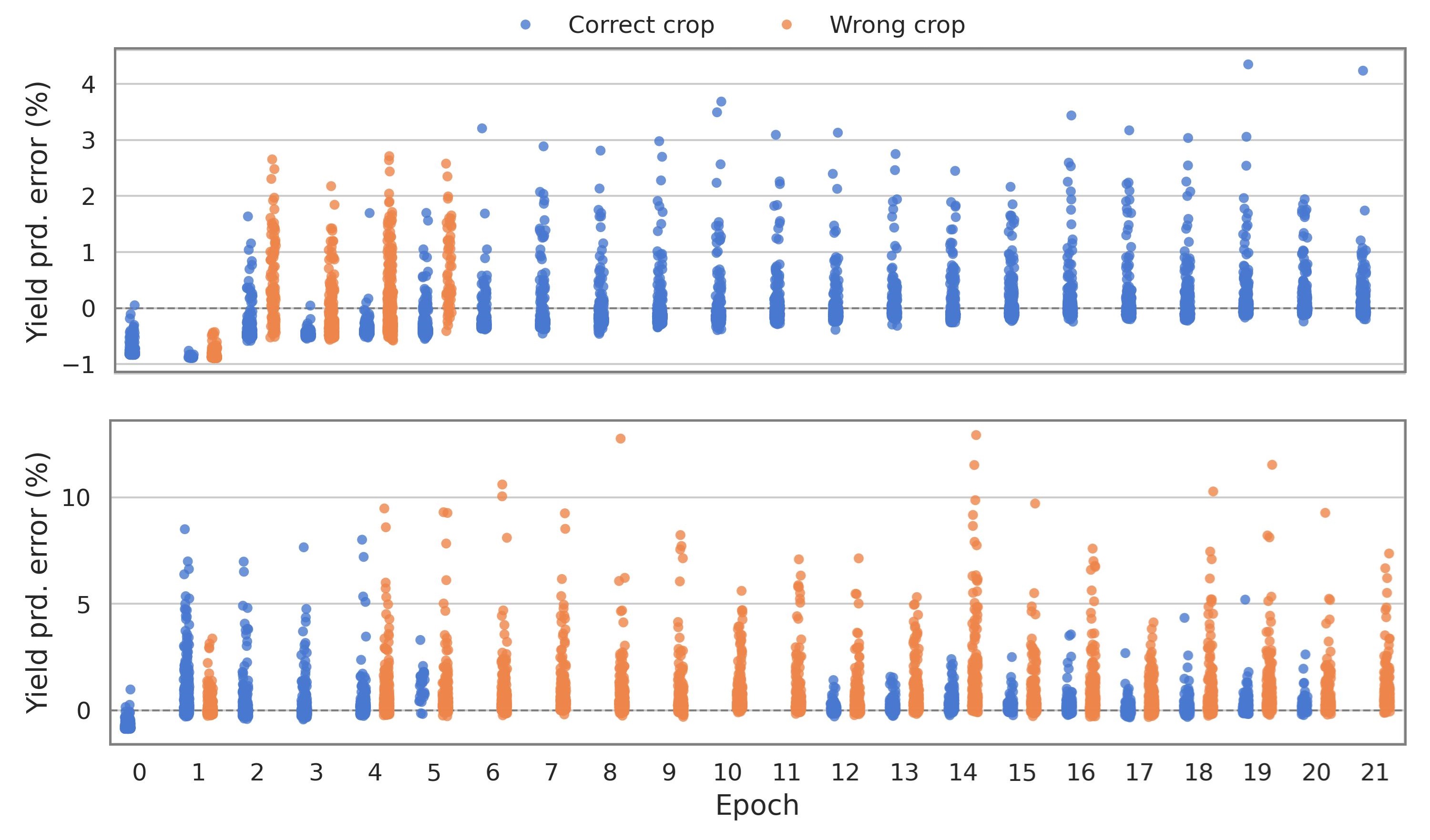}
            \caption{Comparison of model performance on the tasks of yield prediction (measured in relative error) and crop prediction accuracy for the \yc dataset across 21 learning epochs. Results correspond to two soybean fields. 300 correctly classified and another 300 misclassified pixels are displayed for each field.}
            \label{fig:yc_xai_metrics_2flds}
        \end{figure}

\begin{figure}[h]
            \centering
            \includegraphics[width=\textwidth]{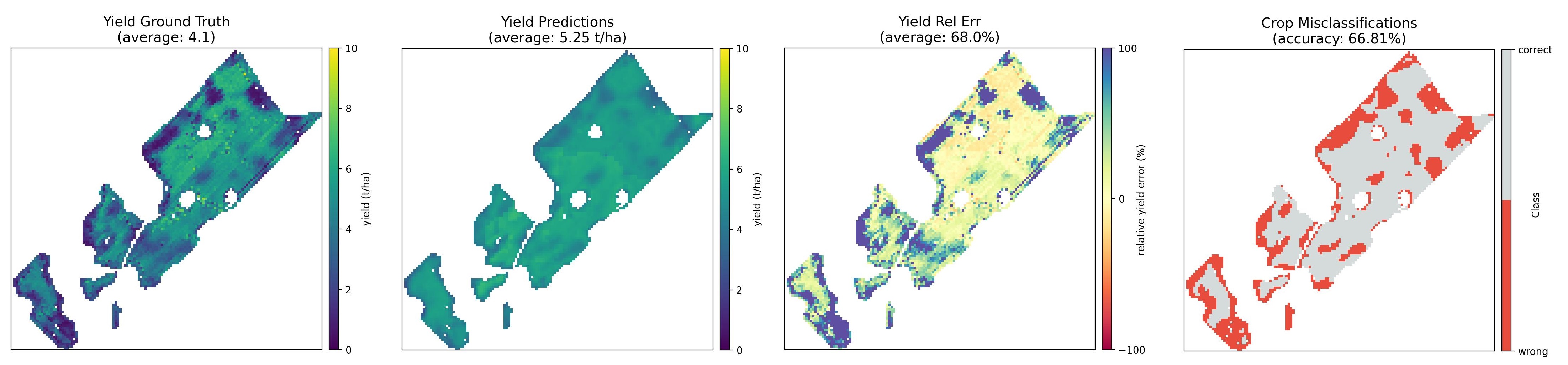}
            \caption{\yc model performance on a soybean field at epoch 16. From left to right: Target yield, predicted yield, relative yield error, and crop misclassifications. More in Appendix~\ref{app:yc_xai}}
            \label{fig:yc_map1}
        \end{figure}

        In the \yc dataset, we evaluate the model performance in Experiment \textbf{7} across epochs, analyzing the relationship between yield relative error and crop prediction accuracy. 
        In a preliminary analysis, we analyze crop-specific performance and include the results in Appendix~\ref{app:yc_xai}. 
        Since a more significant correlation between correct crop classification and improved yield prediction was noticed in soybean fields, we further examine subfield-level performance of two soybean fields by analyzing a random sample of pixels. The results displayed in Figure~\ref{fig:yc_xai_metrics_2flds} show the yield prediction relative error for correctly and incorrectly classified pixels throughout the training. 
        In both fields, the yield prediction relative error is generally higher for misclassified pixels (orange) compared to correctly classified ones (blue), with this effect appearing in early epochs for one field and persisting after the model reaches optimal performance (epoch 11) in another. These findings suggest that incorrect crop classification at the subfield level negatively impacts yield prediction. Additionally, Figure~\ref{fig:yc_map1} illustrates yield and crop type prediction maps at different epochs, from a field where we clearly notice that regions with crop misclassification correspond to areas with significant yield underestimation. More similar examples are included in Appendix~\ref{app:yc_xai}.

        % FB_237
        % LN_112

        \begin{figure}[t]
            \centering
            \includegraphics[width=0.9\linewidth]{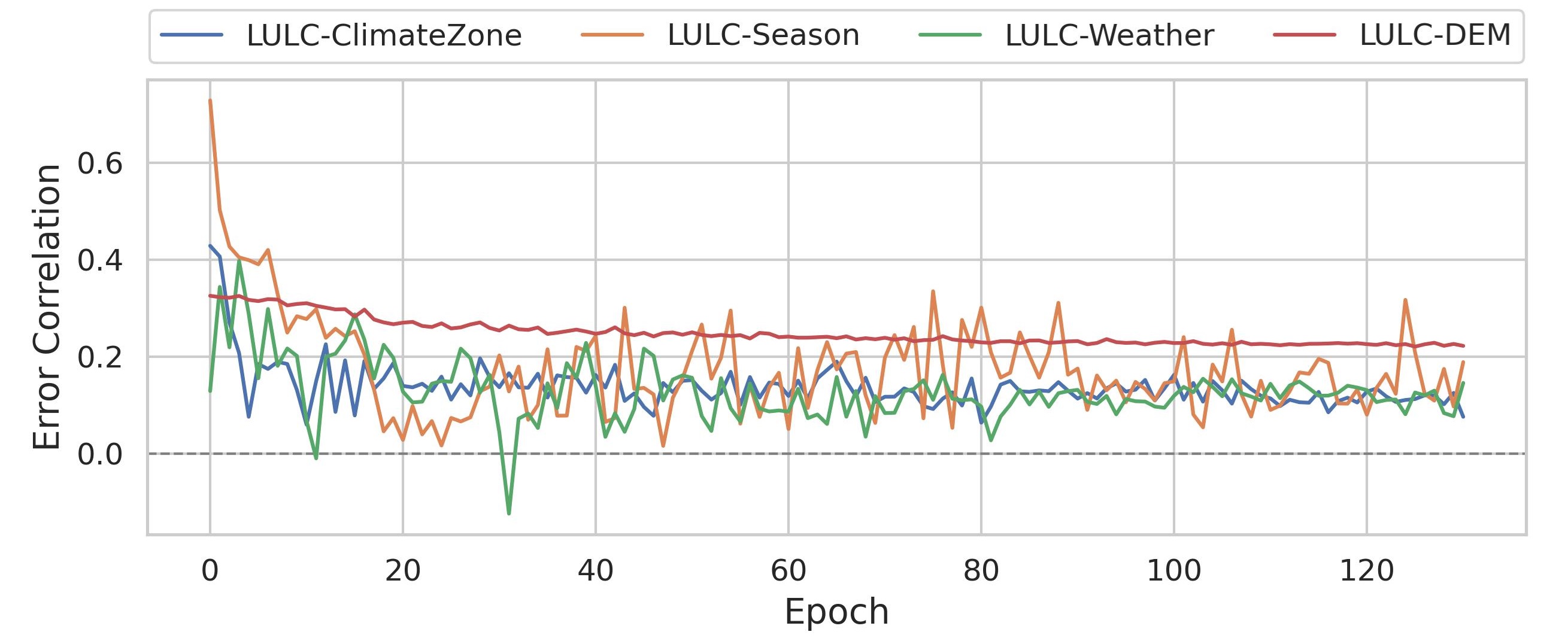}
            \caption{Error correlation between the main \benge task (i.e. \gls{lulc}) and auxiliary tasks.}
            \label{fig:benge_corr}
        \end{figure}

        \begin{figure}[t]
            \centering
            \includegraphics[width=0.8\linewidth]{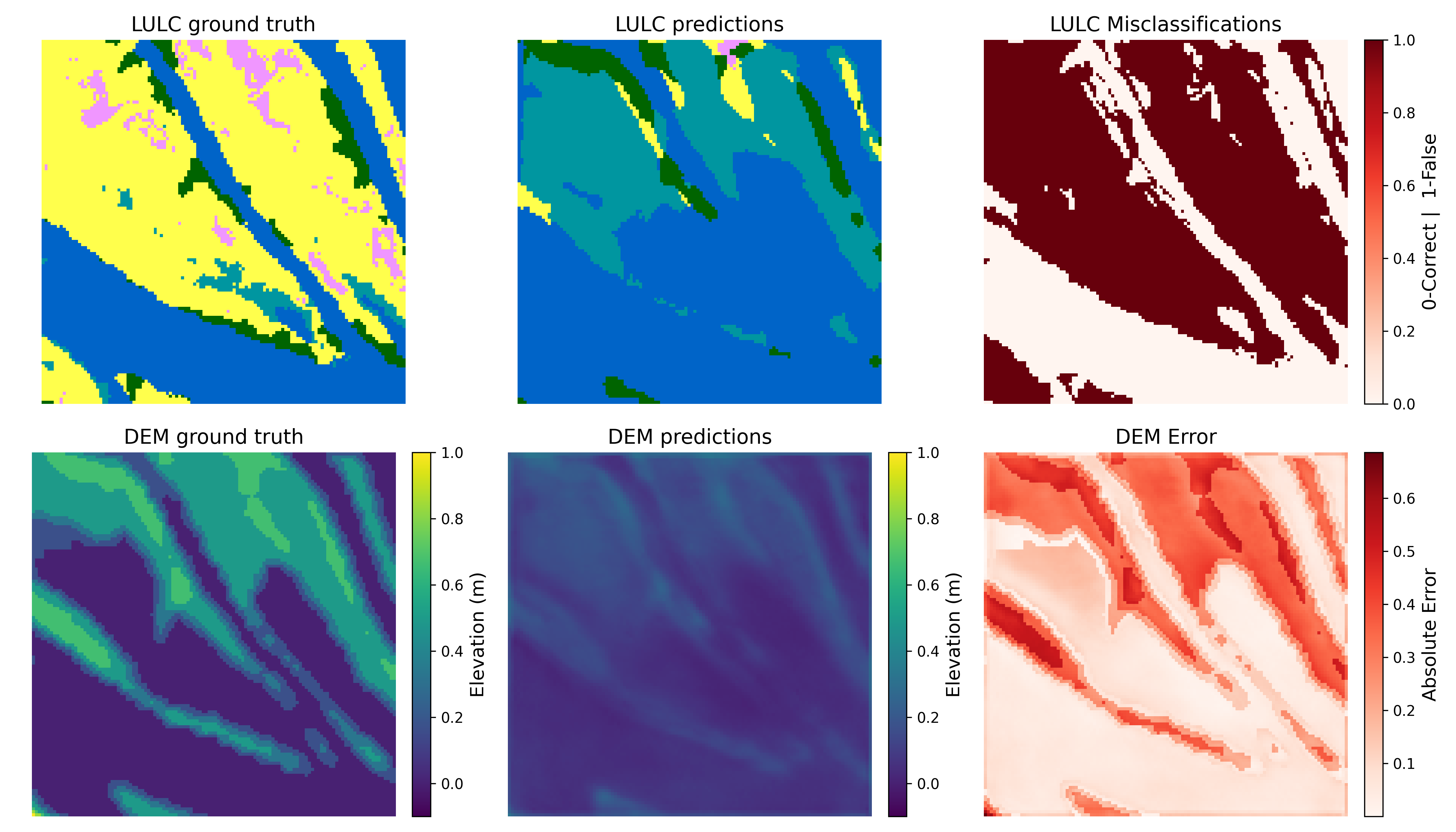}
            \caption{Model predictions and errors, compared against the ground truths, on the \gls{lulc} and \gls{dem} prediction tasks. The predictions are of the best epoch, on a random \benge dataset sample from the test set. More examples in Appendix~\ref{app:benge_xai}}
            \label{fig:lucl_dem_1}
        \end{figure}
        
        \subsubsection{\benge}
        To investigate the explanatory potential of auxiliary tasks in the \benge dataset, we analyze Experiment \textbf{15} (see Table~\ref{tab:benge_res}), which predicts all available modalities as auxiliary tasks.
        We first compute the Pearson correlation between the error of the main task, \gls{lulc} classification, and the errors of the auxiliary tasks, on 10\% of the test set.
        The results presented in Figure~\ref{fig:benge_corr} indicate a decreasing correlation for all task combinations during early training epochs. While the \gls{lulc}-Season correlation exhibits fluctuations throughout training, these variations are less pronounced in the \gls{lulc}-Weather and \gls{lulc}-ClimateZone combinations. 
        In contrast, the \gls{lulc}-\gls{dem} correlation remains more stable, likely due to the similar spatial resolution of both tasks, as they each produce a single-channel image as output. This differs from the other auxiliary tasks, which predict tabular data. Although the correlations do not exceed 0.23, we verified that the p-values remain below 0.05.
        To further examine this correlation between \gls{lulc} and \gls{dem}, we present in Figure~\ref{fig:lucl_dem_1} a data sample where this relationship is clearly visible, with additional examples provided in Appendix~\ref{app:benge_xai}.
        
        Through the examination of a group of samples, we extracted more insightful conclusions regarding the model behavior across tasks; we observed that prediction errors of \gls{lulc} and \gls{dem} tend to correlate in regions where the model fails to accurately determine elevation, particularly along boundaries such as terrain edges or riverbanks. In these regions, land cover classification errors were more frequent. 
        Conversely, when \gls{lulc} misclassifications are scattered within a patch containing highly heterogeneous land cover, the correlation is weak. These areas typically feature stable terrain elevation, leading to \gls{dem} prediction errors that do not exhibit the same scattered distribution.

        \begin{figure}[h]
            \centering
            \includegraphics[width=\linewidth]{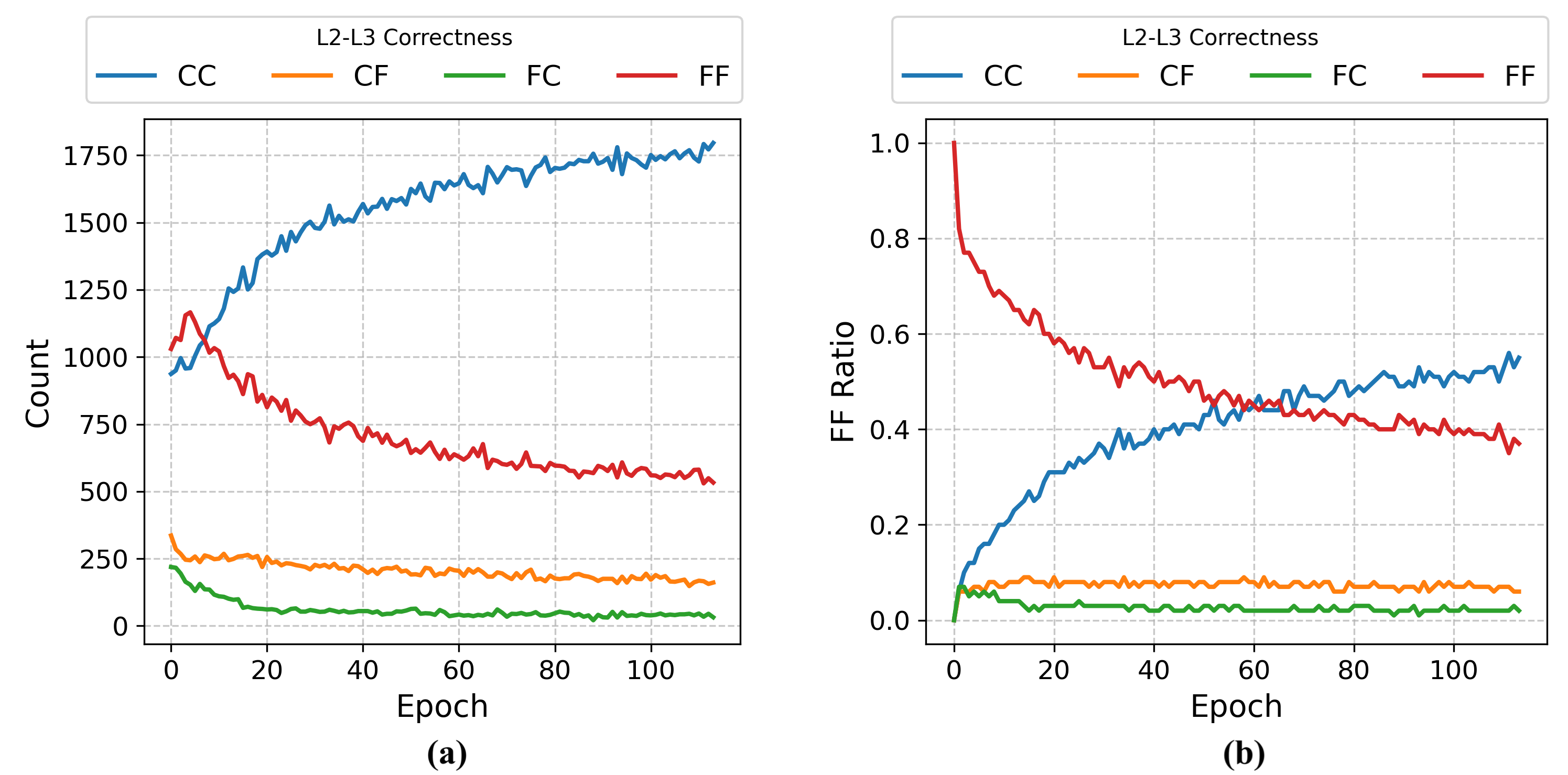}
            \caption{(a) Count of combinations of correct or false classifications of L2 and L3 labels in the test set of \treesat dataset, throughout the training.
            (b) Categorisation ratio of the samples categorized as FF at epoch 0 throughout the training.}
            \label{fig:treesat_countlineplot}
        \end{figure}
        
        \begin{figure}[h]
            \centering
            \includegraphics[width=0.8\linewidth]{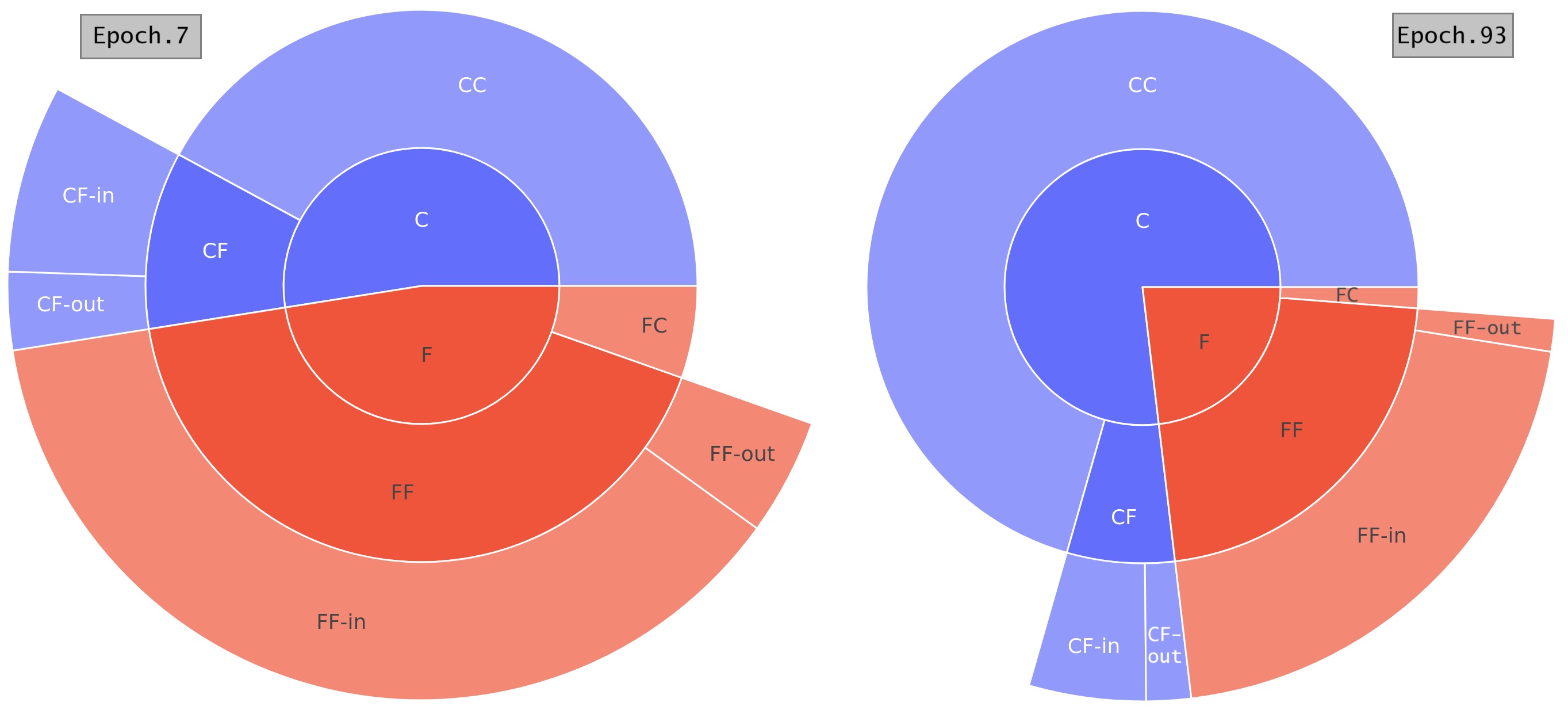}
            \caption{Pie Chart of the distribution of combinations of correct or false classifications of L2 and L3 labels. The results are shown for the test set, inferred at epochs 7 and 93.}
            \label{fig:treesat_pichart}
        \end{figure}
        
        \subsubsection{\treesat}

        We investigate Experiment \textbf{7} in \treesat dataset, which predicts L2 and age alongside the main L3 label. We examine the combinations of L2 and L3 predictions in the test set throughout training, with results presented in Figure~\ref{fig:treesat_countlineplot}.a. 
        Here, 'C' denotes a correctly predicted label, while 'F' indicates a false prediction. The notation follows the order of L2 and L3 predictions; for instance, 'CF' means that L2 was correctly predicted, but L3 was not.
        The results reveal that the count of instances where one label is correct while the other is incorrect (i.e., CF and FC) remain relatively stable throughout training. In contrast, the number of samples where both labels are correct (CC) consistently increases, while instances where both labels are misclassified (FF) decrease correspondingly.
        This trend reveals an interesting pattern about the model behavior; it suggests that FF samples are more likely to be corrected into CC as training progresses, whereas instances in which only one label is initially correct (CF or FC) are less likely to be fully corrected later during the learning process. This hypothesis is verified and confirmed in Figure~\ref{fig:treesat_countlineplot}.b.

        Given the hierarchical nature of tree classes, we further examine how this structure influences the model's predictions. Figure~\ref{fig:treesat_pichart} illustrates the distribution of L2-L3 prediction combinations and their adherence to the hierarchy at an early training epoch (epoch 7) and at the best-performing epoch (epoch 93). 
        We add '-in' to the label of samples where the predicted L3 belongs to the predicted parent class L2 and '-out' to instances where it does not.
        The results indicate that when L3 is misclassified (i.e., in CF and FF cases), the proportion of instances where the predicted L3 remains within the predicted L2 class is consistently higher than those where it falls outside, regardless of whether L2 is correctly predicted. In other words, at both early training stages and the model's peak performance, CF-in is more frequent than CF-out, and FF-in is more frequent than FF-out. This suggests that the model has learned aspects of the hierarchical relationship between L2 and L3 and tends to respect it even when misclassifying L3. Note that in CC cases the hierarchy is always maintained, whereas in FC cases it is always violated. 
        Since the experiment explained here also predicts the age, we include an analysis of the correlation between this modality and different L2-L3 correctness combinations in Appendix~\ref{app:treesat_xai}.

\section{Discussion}\label{sec:discuss}

    While our findings demonstrate the potential of multitask learning for model interpretability, we would like to highlight certain limitations which are to be addressed in future work.
    
    The correlation patterns identified through the analysis of error maps in \benge and \yc were observed in a limited number of samples. However, the presented examples provide evidence of the tight interaction of the model behavior across multiple tasks, and leveraging these observations to correct model errors would enhance performance in both tasks.
    For instance, integrating interpretability insights as constraints within the loss function could enforce meaningful relationships between tasks. Using the hierarchical structure of labels in the \treesat dataset to refine predictions is one example.
    Another promising direction is to refine the selection of task weights in multitask learning. Automating this process using uncertainty estimation \cite{kendall2018multi} or adaptive weighting based on loss improvement rates \cite{liu2019end} could enhance the balance between tasks. We conducted initial experiments to test both approaches, but they were not more successful than the manual selection of weights, yet further experiments are needed. 
    Finally, automating the neural architecture search could further optimize our approach, reducing reliance on manual expertise and improving model performance to align with findings from prior studies in which multitask learning outperformed single-task baselines \cite{maniscalco2024multimodal,ding2019effectiveness,lu202012,kendall2018multi,sener2018multi,liu2019end,levering2021relation}.

\section{Conclusion}\label{sec:conclusion}

    In this work, we proposed a multitask learning framework to enhance model explainability in \gls{rs}. We exploited the rich information content of satellite data to shift additional input modalities into auxiliary tasks. 
    This approach not only maintained comparable performance to baseline models but also reduced the need for additional data at deployment. More importantly, it provided valuable explainability insights through the analysis of error correlations between the main and auxiliary tasks across three diverse \gls{rs} datasets. We demonstrate how this analysis can improve understanding of model reasoning and inner workings. Further, focusing on a specific use case and conducting deeper analysis could yield even greater insights into the model behavior.
    Future work will integrate these insights into the data preparation and modeling pipeline to refine model reasoning and, consequently, enhance performance. 

\newpage
\section*{Appendix}

% ---- Appendix ----
\appendix

\section{Datasets and Modalities}\label{app:data}
    % provide more details about YC
    % include table
    % specify which modality is normalized or the range
    % clarify how the weather variable in BenGe were fewer in MTL
        
        In the \yc dataset, yield maps were collected by combine harvesters at harvesting date for three crop types, across multiple fields in Argentina, as summarized in Table~\ref{tab:yc_data}.
        The harvester records equidistant data points at a high spatial resolution, including information about the yield in tons per hectare (t/ha).
        Yield points were rasterized, by averaging all yield points that fall within the 10x10m grid cell matching the spatial resolution of the corresponding satellite images. 
        These are collected from the Sentinel-2 Level-2A satellite mission, including all available scenes from seeding to harvesting dates. The images are multispectral, including 12 spectral bands, to which we add 13 bands corresponding to the \gls{scl} labels. To match the resolution across channels and facilitate the pixel-wise processing of the \yc dataset, spectral bands with lower resolutions are upsampled to 10m resolution.
        Within the spatial boundaries of each field, we collect two additional modalities, including weather data derived from the ECMWF Reanalysis (ERA5) \cite{hersbach2020era5} in 30km resolution, and \gls{dem} data from NASA’s \gls{srtm} \cite{farr2000shuttle} in 30m resolution.
            Weather data is aggregated for each day at field level for minimum, maximum, and mean temperature and total precipitations.
            For DEM, in addition to the elevation values, we derived the aspect, curvature, slope and the \gls{twi}.
            Soil and DEM data were transformed into raster images and upsampled to a 10m resolution, using a cubic spline interpolation.

        \begin{table}[]
            \centering
            \caption{\yc dataset description.}
            \label{tab:yc_data}
            \begin{tabular}{lcccccccccl}
                \hline
                \textbf{\quad} & \textbf{Crop} & \textbf{\quad} & \textbf{\# Farms} & \textbf{\quad} & \textbf{\# Fields} & \textbf{\quad} & \textbf{\# Pixels} & \textbf{\quad} & \textbf{Percentage} &   \textbf{\quad}  \\ \cline{2-10}
                 & Corn &  & 21 &  & 147 &  & 1,003,133 &  & 27.8\% &  \\
                 & Soybean &  & 29 &  & 289 &  & 2,103,250 &  & 58.4\% &  \\
                 & Wheat &  & 13 &  & 61 &  & 497,651 &  & 13.8\% &  \\ \cline{2-10}
                 & Total &  & 63 &  & 497 &  & 3,604,034 &  & 100\% &  \\ \hline
            \end{tabular}
        \end{table}

        The descriptions for \benge and \treesat datasets are detailed in \cite{mommert2023ben} and \cite{ahlswede2022treesatai}, respectively. 
        For weather data in \benge, we include all five weather features (i.e. temperature, two wind vectors, relative humidity, and atmospheric pressure) when the modality is used as input data. However, when utilizing weather data as an auxiliary target, we exclude the wind vectors.
        Table~\ref{tab:modalities} provides a summary of the modalities used in each dataset, highlighting the main input modalities and the main target. Additionally, the table specifies the type of input encoder used for modalities when implemented as input data, as well as the type of prediction heads, loss functions, and evaluation metrics applied when modalities are used as targets.

        \begin{table}[]
        \centering
        \caption{Available and used data modalities in the three multimodal datasets: \yc, \benge, and \treesat. The main input and target modalities are highlighted in bold.}
        \adjustbox{center=\textwidth}{
        \begin{tabular}{ccccccccccccccc}
        \hline
        \textbf{ \quad } & \textbf{Dataset} & \textbf{} & \textbf{Modality} & \textbf{} & \textbf{Type} & \textbf{} & \textbf{Encoder} & \textbf{} & \textbf{Prediction Head} & \textbf{} & \textbf{Loss function} & \textbf{} & \textbf{Metric} &  \textbf{ \quad }  \\ \cline{2-14}
         & \multirow{5}{*}{\yc} &  & \textbf{Sat (S2)} &  & TS of 25 features &  & Transformer &  & - &  & - &  & - &  \\
         &  &  & \textbf{Weather} &  & TS of 4 features &  & Transformer &  & - &  & - &  & - &  \\
         &  &  & \textbf{Yield} &  & single scalar &  & - &  & Reg. MLP &  & MSE &  & $\text{R}^2$ &  \\
         &  &  & Crop label &  & 3 classes &  & MLP &  & Class. MLP &  & Cross entropy &  & micro-F1 &  \\
         &  &  & \gls{dem} &  & 5 features &  & MLP &  & Reg. MLP &  & MSE &  & MAE &  \\ \cline{2-14}
         & \multirow{6}{*}{\benge} &  & \textbf{Sat (S1,S2)} &  & multichannel image &  & U-Net &  & - &  & - &  & - &  \\
         &  &  & \textbf{LULC} &  & segmentation mask &  & - &  & Multiclass Segmentation &  & Cross entropy &  & IoU &  \\
         &  &  & Elevation &  & single channel image &  & U-Net &  & Dense Segmentation &  & MSE &  & MAE &  \\
         &  &  & Climate Zone &  & 12 classes &  & Embeddings &  & Class. MLP &  & Cross entropy &  & micro-F1 &  \\
         &  &  & Season &  & single scalar &  & MLP &  & Reg. MLP &  & MSE &  & MAE &  \\
         &  &  & Weather &  & 5 features &  & MLP &  & Reg. MLP &  & MSE &  & MAE &  \\ \cline{2-14}
         & \multirow{6}{*}{\treesat} &  & \textbf{Aerial} &  & multichannel image &  & CNN &  & - &  & - &  & - &  \\
         &  &  & \textbf{Sat (S1,S2)} &  & multichannel image &  & MLP &  & - &  & - &  & - &  \\
         &  &  & \textbf{Level-3 (L3)} &  & 15 classes &  & - &  & Class. MLP &  & Cross entropy &  & micro-F1 &  \\
         &  &  & Level-2 (L2) &  & 9 classes &  & - &  & Class. MLP &  & Cross entropy &  & micro-F1 &  \\
         &  &  & Level-1 (L1) &  & 3 classes &  & - &  & Class. MLP &  & Cross entropy &  & micro-F1 &  \\
         &  &  & Age &  & single scalar &  & MLP &  & Reg. MLP &  & MSE &  & MAE &  \\ \hline
        \end{tabular}
        }
        \adjustbox{center=\textwidth}{\parbox{1.5\textwidth}{\center \small Reg.: Regression | TS: Time Series | Sat: Satellite | S1: Sentinel-1 | S2: Sentinel-2  }}
        \label{tab:modalities}
        \end{table}

        All the three datasets have been split into training, validation, and test sets.
        In \yc dataset, 60\% is used for training, 20\% validation, and 20\% for testing.
        Since each input sample represents a pixel from a field, we grouped samples by field before splitting the data, to ensure that the model encounters unseen fields in the validation and test splits. To maintain a consistent data distribution, we stratified the splits by year, ensuring that each split contains data from all years.
        In \benge dataset, we use the 80/10/10 split provided by the dataset authors.
        In \treesat, we use the 90/10 split provided for training and testing, and further split the 10\% into validation and testing sets.

\section{Multimodal and Multitask Models}
    For the modeling stage, an overview of the loss functions and evaluation metrics used per dataset and modality are included in Table~\ref{tab:modalities}.
    We further include and describe in this section additional experiments conducted in the \yc and \benge datasets.
    
    As we evaluated various network configurations across different datasets on their respective validation set, we explored diverse architectural types, adjusting the number of layers, the hidden layer sizes, data sampling strategies, and the loss weights. Hence, we describe below for each dataset the architectural configurations of the experiments used in the explanatory analysis from Section~\ref{ssec:xai}.

    \subsection{\yc}\label{app:yc_mod}
        
        \begin{table}[t]
        \caption{Modeling performance on the test set of the \yc dataset. The best and second-best scores are highlighted in bold and underlined, respectively. Crop classification performance is given in micro F1 score.}
        \centering
        \adjustbox{center=\textwidth}{
        \begin{tabular}{ccccccccccccccccc}
            \hline
             & \multicolumn{2}{c}{} &  & \multicolumn{4}{c}{\textbf{Modalities}} &  & \multicolumn{4}{c}{\textbf{Main task}} &  & \multicolumn{2}{c}{\textbf{Auxiliary tasks}} &  \\ \cline{5-8} \cline{10-13} \cline{15-16}
            \quad \textbf{ } & \multicolumn{2}{c}{\textbf{Experiment}} & \quad \textbf{ } & \textbf{ Satellite } & \textbf{\begin{tabular}[c]{@{}c@{}} Crop\\ label \end{tabular}} & \textbf{ Weather } & \textbf{ DEM } & \quad \textbf{ } & \textbf{\begin{tabular}[c]{@{}c@{}}Yield\\ (R²)\end{tabular}} & \textbf{\begin{tabular}[c]{@{}c@{}}Yield\\ (R²-Soybean)\end{tabular}} & \textbf{\begin{tabular}[c]{@{}c@{}}Yield\\ (R²-Wheat)\end{tabular}} & \textbf{\begin{tabular}[c]{@{}c@{}}Yield\\ (R²-Corn)\end{tabular}} & \quad \textbf{ } & \textbf{\begin{tabular}[c]{@{}c@{}}Crop cls.\\ (F1)\end{tabular}} & \textbf{\begin{tabular}[c]{@{}c@{}}DEM\\ (MAE)\end{tabular}} & \quad \textbf{ } \\ \cline{2-3} \cline{5-8} \cline{10-13} \cline{15-16}
             & Baselines & 1.a &  & \inmod & (soybean) &  &  &  & 0.64 & \textbf{0.64} & - & - &  & - & - &  \\
             &  & 1.b &  & \inmod & (wheat) &  &  &  & 0.64 & - & 0.64 & - &  & - & - &  \\
             &  & 1.c &  & \inmod & (corn) &  &  &  & 0.48 & - & - & 0.48 &  & - & - &  \\
             &  & 1.d &  & \inmod & (all crops) &  &  &  & {\ul 0.81} & 0.45 & 0.79 & {\ul 0.62} &  & - & - &  \\ \cline{2-3} \cline{5-8} \cline{10-13} \cline{15-16}
             & MML & 2 &  & \inmod & \inmod &  &  &  & 0.77 & 0.44 & 0.78 & 0.51 &  & - & - &  \\
             &  & 3 &  & \inmod &  & \inmod &  &  & 0.75 & 0.37 & {\ul 0.80} & 0.45 &  & - & - &  \\
             &  & 4 &  & \inmod & \inmod & \inmod &  &  & 0.79 & 0.40 & 0.78 & 0.59 &  & - & - &  \\
             &  & 5 &  & \inmod &  & \inmod & \inmod &  & {\ul 0.81} & 0.45 & 0.78 & \textbf{0.63} &  & - & - &  \\
             &  & 6 &  & \inmod & \inmod & \inmod & \inmod &  & 0.79 & 0.42 & 0.75 & 0.57 &  & - & - &  \\ \cline{2-3} \cline{5-8} \cline{10-13} \cline{15-16}
             & MTL & 7 &  & \inmod & \outmod &  &  &  & \textbf{0.82} & {\ul 0.52} & \textbf{0.82} & \textbf{0.63} &  & {\ul 99.4} & - &  \\
             &  & 8 &  & \inmod & \outmod & \inmod &  &  & 0.77 & 0.48 & 0.77 & 0.49 &  & \textbf{99.5} & - &  \\
             &  & 9 &  & \inmod & \outmod & \inmod & \inmod &  & 0.80 & 0.43 & 0.75 & 0.60 &  & \textbf{99.5} & - &  \\
             &  & 10 &  & \inmod & \outmod & \inmod & \outmod &  & 0.75 & 0.37 & 0.78 & 0.48 &  & 99.3 & \textbf{0.42} &  \\ \hline
        \end{tabular}
        }
        \adjustbox{center=\textwidth}{\parbox{1.5\textwidth}{\centering \small   \inmod \textbf{} Input | \outmod \textbf{} Output \quad | \quad MML:Multimodal learning \quad | \quad MTL:Multitask Learning \quad |\quad cls.:classification.}}
        \label{tab:yc_res_full}
        \end{table}

        \paragraph{Additional Experiments:} In Table~\ref{tab:yc_res_full} we extend the baseline experiments to analyze the model performance per crop-type.
            The first three experiments (\textbf{1.a} - \textbf{1.c}) train the model using satellite data alone, based on the subset data of each crop individually, while all subsequent experiments merge samples from all crops types.
            The first four baseline experiments indicate that combining crop types has a positive impact on the overall model performance, achieving the relatively high \gls{r2} score of 0.81.
            Evaluating the performance per crop type reveals an increase of 0.15 and 0.04 in the \gls{r2} score of wheat and corn pixels, respectively. Nevertheless, a notable decline of 0.19 is observed for soybean fields. 
            Despite using weighted data sampling during the training to mitigate class imbalance, these results correlate with the size of each crop type within the dataset, as we observe that the smallest crop subset (wheat) benefits the most, followed by the second smallest (corn). In contrast, the largest soybean dataset exhibited a decline, and performed better when trained individually, in Experiment \textbf{1.a}. As a result, corn and wheat samples benefit from the data mixing, unlike soybean samples.
            The gap observed between the global vs. crop-specific \gls{r2} scores is caused by the nature of this score, and the gap confirms that the model's performance is not consistent across different crop types.
        
        \paragraph{Analyzed Experiment:}
        Experiment \textbf{7} for \yc dataset processes the satellite modality pixel-wise (time series of 25 channels, 12 for the spectral bands and 13 for the scene classification mask label) using a Transformer-based architecture with single attention head and 4 layers, and uses the number of days to harvest for positional encodings \cite{vaswani2017attention}. The regression head for yield prediction consists of a two fully connected layers with BatchNorm and ReLU, mapping the 32-dimensional features return by the satellite encoder to a single output. The crop classification head follows a similar structure but maps features to 3 classes.
        In the total loss, the yield prediction task is assigned a weight of 0.67, while the crop classification task is assigned a weight of 0.33. 
        
    \subsection{\benge}\label{app:benge_mod}

        \begin{table}[t]
        \caption{Test set performance on the \benge dataset. The best and second-best scores are highlighted in bold and underlined, respectively. Climate zone classification performance is given in micro F1 score.}
        \centering
        \adjustbox{center=\textwidth}{
        \begin{tabular}{cccccccccccccccccc}
        \hline
         & \multicolumn{2}{c}{} &  & \multicolumn{5}{c}{\textbf{Modalities}} &  & \multicolumn{2}{c}{\textbf{Main task}} &  & \multicolumn{4}{c}{\textbf{Auxiliary tasks}} &  \\ \cline{5-9} \cline{11-12} \cline{14-17}
        \quad \textbf{  } & \multicolumn{2}{c}{\textbf{Experiment}} & \quad \textbf{ } & \textbf{ Satellite } & \textbf{ Elevation } & \textbf{\begin{tabular}[c]{@{}c@{}} Climate\\ Zone \end{tabular}} & \textbf{ Season } & \textbf{ Weather } & \quad \textbf{ } & \textbf{\begin{tabular}[c]{@{}c@{}}LULC\\ (Accuracy)\end{tabular}} & \textbf{\begin{tabular}[c]{@{}c@{}}LULC\\ (IoU)\end{tabular}} & \quad \textbf{ } & \textbf{\begin{tabular}[c]{@{}c@{}}Elevation\\ (MAE)\end{tabular}} & \textbf{\begin{tabular}[c]{@{}c@{}}Climate zone \\ (F1)\end{tabular}} & \textbf{\begin{tabular}[c]{@{}c@{}}Season\\ (MAE)\end{tabular}} & \textbf{\begin{tabular}[c]{@{}c@{}}Weather\\ (MAE)\end{tabular}} & \quad \textbf{ } \\ \cline{2-3} \cline{5-9} \cline{11-12} \cline{14-17}
         & Baseline & 1 &  & \inmod &  &  &  &  &  & {\ul 87.94} & {\ul 0.388} &  & - & - & - & - &  \\ \cline{2-3} \cline{5-9} \cline{11-12} \cline{14-17}
         & MML & 2 &  & \inmod & \inmod &  &  &  &  & 87.91 & 0.386 &  & - & - & - & - &  \\
         &  & 3 &  & \inmod &  & \inmod &  &  &  & 87.90 & 0.386 &  & - & - & - & - &  \\
         &  & 4 &  & \inmod &  &  & \inmod &  &  & 87.91 & \textbf{0.389} &  & - & - & - & - &  \\
         &  & 5 &  & \inmod &  &  &  & \inmod &  & 87.93 & 0.387 &  & - & - & - & - &  \\
         &  & 6 &  & \inmod & \inmod &  & \inmod &  &  & 87.90 & 0.385 &  & - & - & - & - &  \\
         &  & 7 &  & \inmod & \inmod &  &  & \inmod &  & \textbf{87.95} & 0.383 &  & - & - & - & - &  \\
         &  & 8 &  & \inmod & \inmod & \inmod & \inmod & \inmod &  & 87.85 & 0.387 &  & - & - & - & - &  \\ \cline{2-3} \cline{5-9} \cline{11-12} \cline{14-17}
         & MTL & 9 &  & \inmod & \outmod &  &  &  &  & 87.90 & 0.380 &  & {\ul 0.162} & - & - & - &  \\
         &  & 10 &  & \inmod &  & \outmod &  &  &  & 87.93 & 0.379 &  & - & \textbf{94.88} & - & - &  \\
         &  & 11 &  & \inmod &  &  & \outmod &  &  & 87.91 & 0.380 &  & - & - & \textbf{7e-8} & - &  \\
         &  & 12 &  & \inmod &  &  &  & \outmod &  & 87.91 & 0.381 &  & - & - & - & {\ul 0.018} &  \\
         &  & 13 &  & \inmod & \outmod &  & \outmod &  &  & 87.91 & 0.377 &  & {\ul 0.162} & - & {\ul 9e-8} & - &  \\
         &  & 14 &  & \inmod & \outmod &  &  & \outmod &  & 87.89 & 0.377 &  & \textbf{0.161} & - & - & {\ul 0.018} &  \\
         &  & 15 &  & \inmod & \outmod & \outmod & \outmod & \outmod &  & 87.89 & 0.373 &  & {\ul 0.162} & {\ul 94.77} & 5e-5 & \textbf{0.015} &  \\ \hline
        \end{tabular}
        }
        %\adjustbox{center=\textwidth}{\parbox{1.5\textwidth}{\centering \small   \inmod \textbf{} Input | \outmod \textbf{} Output \quad | \quad MML:Multimodal learning \quad | \quad MTL:Multitask Learning \quad | \quad Sat:Satellite \quad | \quad cls.:classification.}}
        %\adjustbox{center=\textwidth}{\parbox{1.5\textwidth}{\centering \small \inmod \textbf{} Input \quad | \quad \outmod \textbf{} Output \quad | \quad  MML:Multimodal Learning \quad | \quad MTL:Multitask Learning \quad | \quad Sat:Satellite images.}}
        \label{tab:benge_res_full}
        \end{table}

        \paragraph{Additional Experiments:}
        In Table~\ref{tab:benge_res_full} we extend the results of \benge modeling experiments by including model performance on the auxiliary tasks.
        We observe that climate zone classification (with 12 classes) achieves a high F1 score close to 95\%. Similarly, the season prediction task yields very low \gls{mae} scores, particularly in comparison to the errors observed in elevation and weather predictions. It is important to note that weather and season data are normalized, whereas elevation values range between 0 and 1.

        \paragraph{Analyzed Experiment:}
        The \benge model from Experiment \textbf{15} explained in Section~\ref{ssec:xai} is a semantic segmentation model for the \gls{lulc} task. It processes two satellite images: 2-channel SAR imagery (from Sentinel-1 mission) and 12-channel multispectral imagery (from Sentinel-2 mission), each encoded using a UNetBackbone with four downsampling and upsampling layers. The modalities are mapped into 64-channel images which are concatenated channel-wise and processed at the fusion block, consisting of two 1x1 convolutional layers with ReLU activations, reducing the combined feature dimension 128 to 64. 
        The model performs multiple tasks, each with a specific prediction head:
        The \gls{lulc} segmentation head uses a simple 1x1 convolutional layer to map the 64-channel fused features to 12 classes, suitable for the multiclass segmentation. The climate zone classification head first applies a 1x1 convolution to reduce features, followed by a fully connected layer with dropout (0.5 probability) to output 12 classes. The season regression head employs a 1x1 convolution to expand features to 16 channels, followed by two fully connected layers with ReLU and dropout (0.2 probability), and a sigmoid activation for bounded regression output. The weather regression head follows a similar structure but outputs three continuous values without activation. Lastly, the \gls{dem} regression head uses two 3x3 convolutional layers, followed by a 1x1 convolution to produce per-pixel elevation values.
        The \gls{lulc} task is assigned a weight of 9, and the remaining four tasks each have a weight of 1. These weights are scaled to ensure they sum to 1.

    \subsection{\treesat}\label{app:treesat_mod}
        \paragraph{Analyzed Experiment:}
        Experiment \textbf{7} explained for the \treesat dataset processes three input modalities: 3-channel SAR imagery (from Sentinel-1 mission), 12-channel multispectral imagery (from Sentinel-2 mission), and 3-channel RGB aerial imagery. Following the recommendations in \cite{ahlswede2022treesatai}, the satellite modalities are flattened and encoded using fully connected networks with three linear layers with ReLU and dropout (0.3 probability) while the aerial images are encoded using a pretrained ResNet18 \cite{he2016deep}, which we fine-tune during the training. 
        The fusion mechanism consists of a simple concatenation of the flattened modality representations. 
        The model subsequently performs three tasks using task-specific heads:
        The L3 head is a simple linear layer that maps the 1536-dimensional fused features to 15 classes. The L2 head uses a two-layer fully connected network with ReLU activation and dropout (0.25 probability) to map the features to nine classes. The age prediction head uses a single linear layer to predict a continuous value.
        The L3 classification task is assigned a weight of 4, and the remaining two tasks each have a weight of 1. These weights are scaled to ensure they sum to 1.

\section{Explainability}

    In this section, we include additional results from the explanatory analysis for each dataset.
    
    \subsection{\yc}\label{app:yc_xai}
    
\begin{figure}
            \centering
            \includegraphics[width=0.95\textwidth]{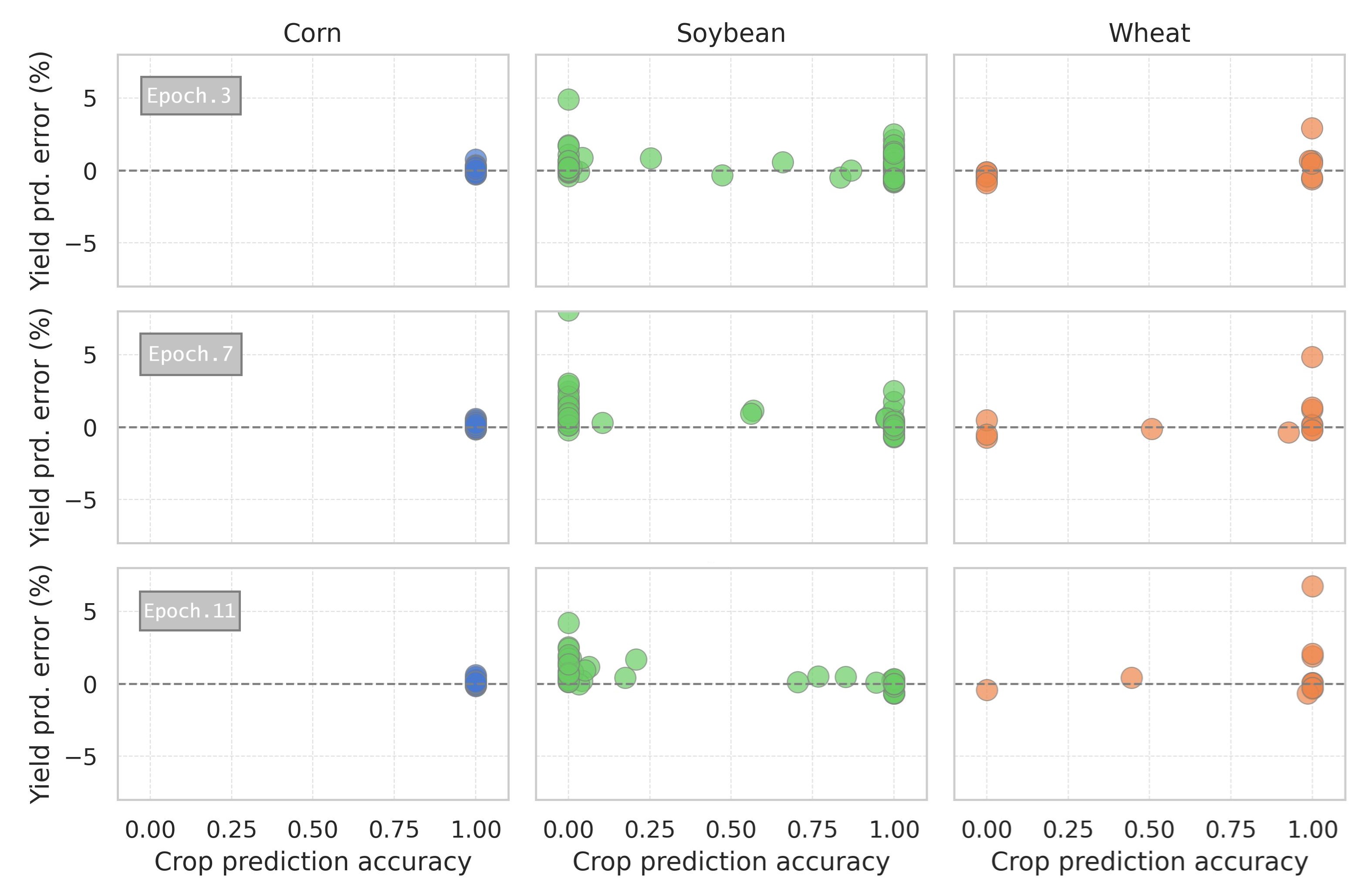}
            \caption{Comparison of model performance on the tasks of yield prediction and crop prediction for the \yc dataset. The rows correspond to the results for epochs 7, 15, and 26, from top to bottom.}
            \label{fig:yc_xai_metrics}
        \end{figure}

        Figure~\ref{fig:yc_xai_metrics} presents results for epochs 3, 7, and 11, separately for each crop, with performance averaged per field. Each point in the figure represents a field, including training, validation, and test sets. We stop in this analysis at epoch 11 as the model achieved its best performance on the validation set at this epoch.
        The figure shows that corn fields consistently exhibit strong yield prediction performance and perfect crop classification accuracy, whereas the model faces greater challenges with the other two crop types. 
        For soybean fields, a decrease in maximum yield prediction error is observed across epochs in fields with high crop classification accuracy, while fields with poor classification maintain high yield prediction errors, suggesting a correlation between correct crop classification and improved yield prediction.
        In wheat fields, fewer instances of poor crop classification are observed as training progresses.
        
\begin{figure}
            \centering
            \includegraphics[width=\textwidth]{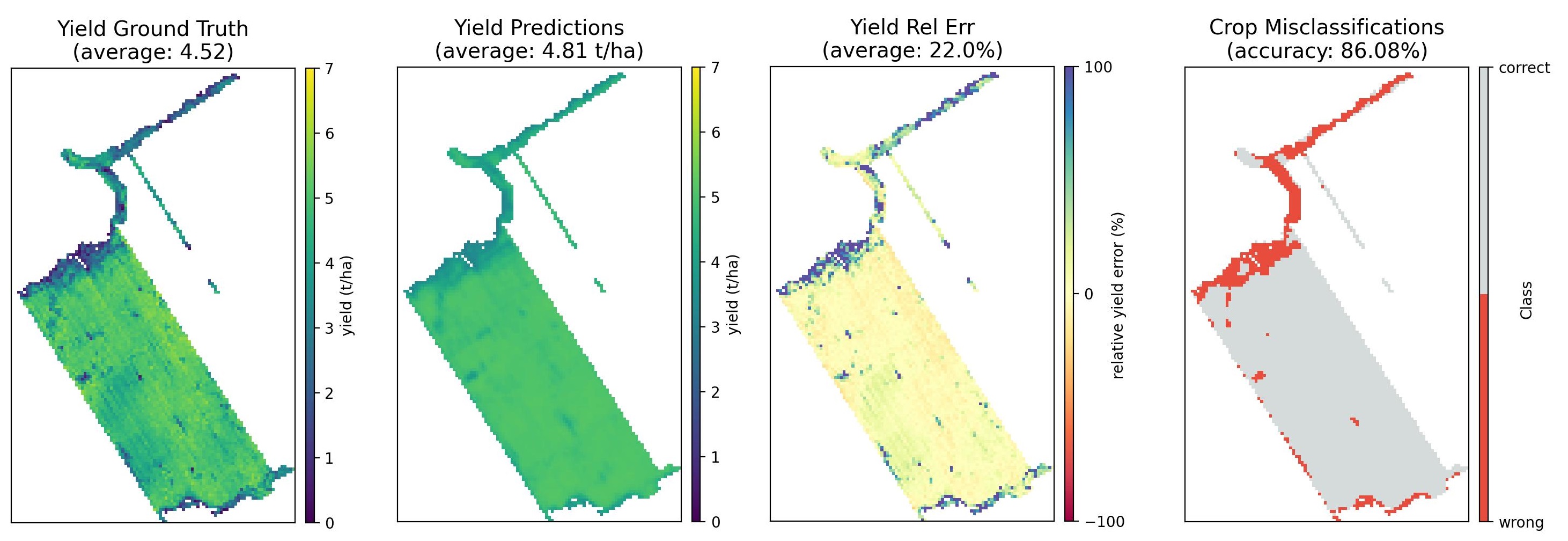}
            \includegraphics[width=\textwidth]{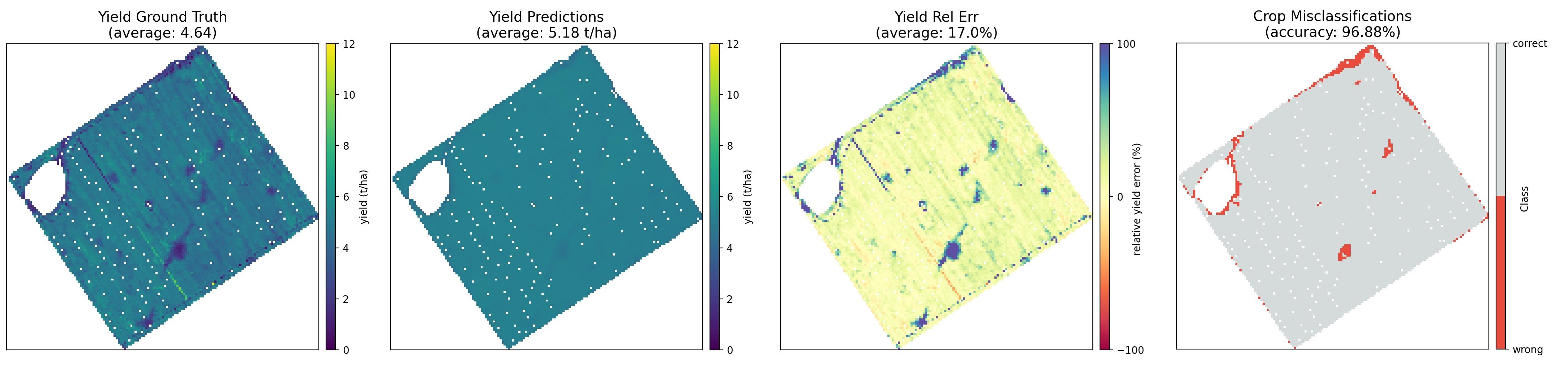}
            \includegraphics[width=\textwidth]{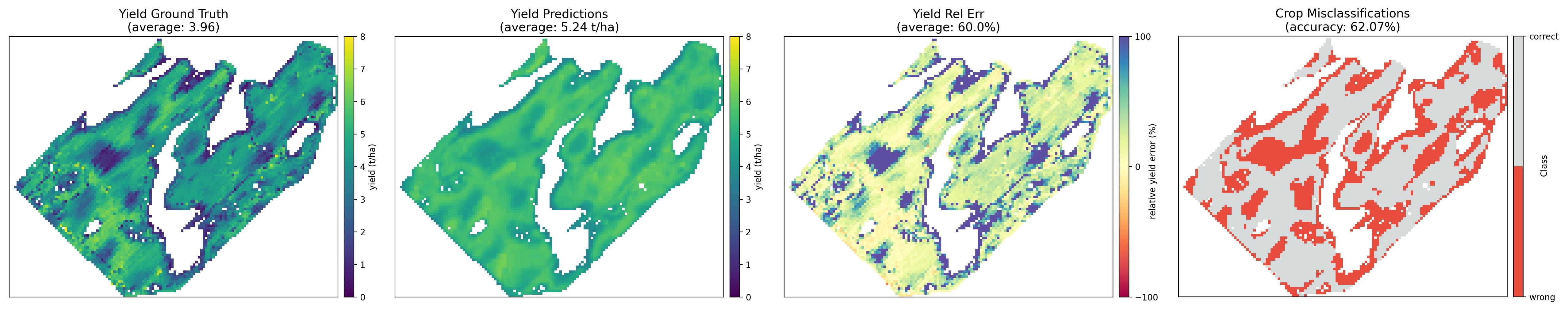}
            \includegraphics[width=\textwidth]{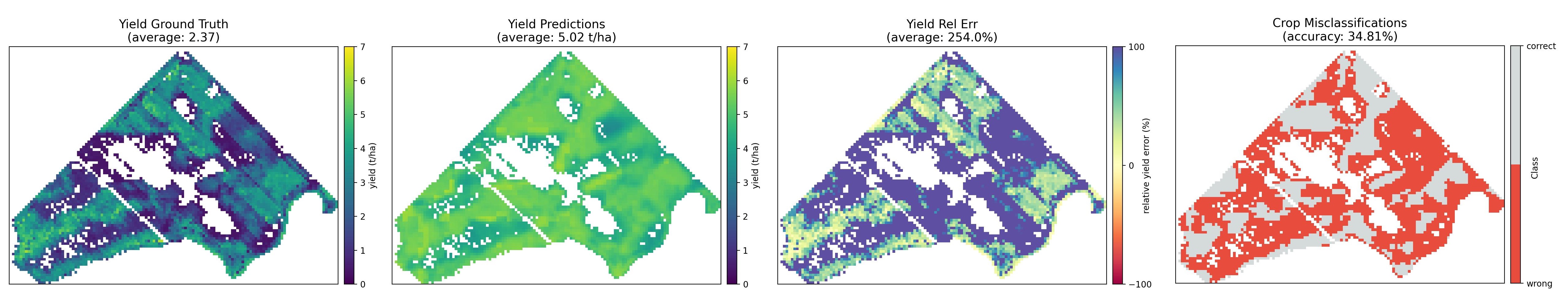}
            \caption{\yc model performance on four soybean fields, at epoch 4 for the two fields at the top and epoch 14 for the two others. From left to right: Target yield, predicted yield, relative yield error, and crop misclassifications.}
            \label{fig:yc_map2}
        \end{figure}

        Figure~\ref{fig:yc_map2}  illustrates yield and crop type prediction maps for different soybean fields and at different epochs, showing that regions with crop misclassification correspond to areas with significant yield underestimation.

    \subsection{\benge}\label{app:benge_xai}
\begin{figure}
            \centering
            \includegraphics[width=0.9\linewidth]{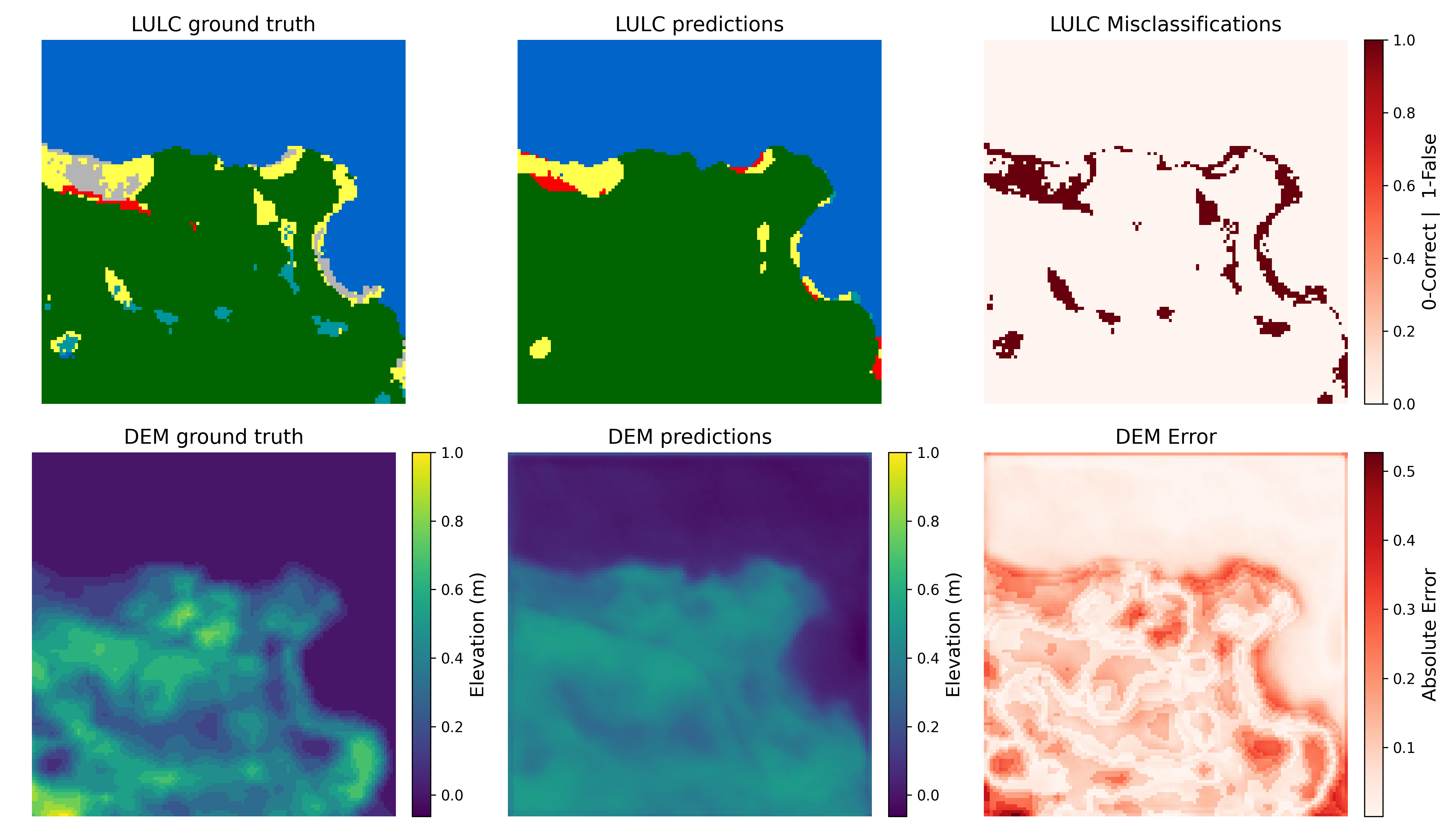}
            \rule{\textwidth}{0.1mm}
            \includegraphics[width=0.9\linewidth]{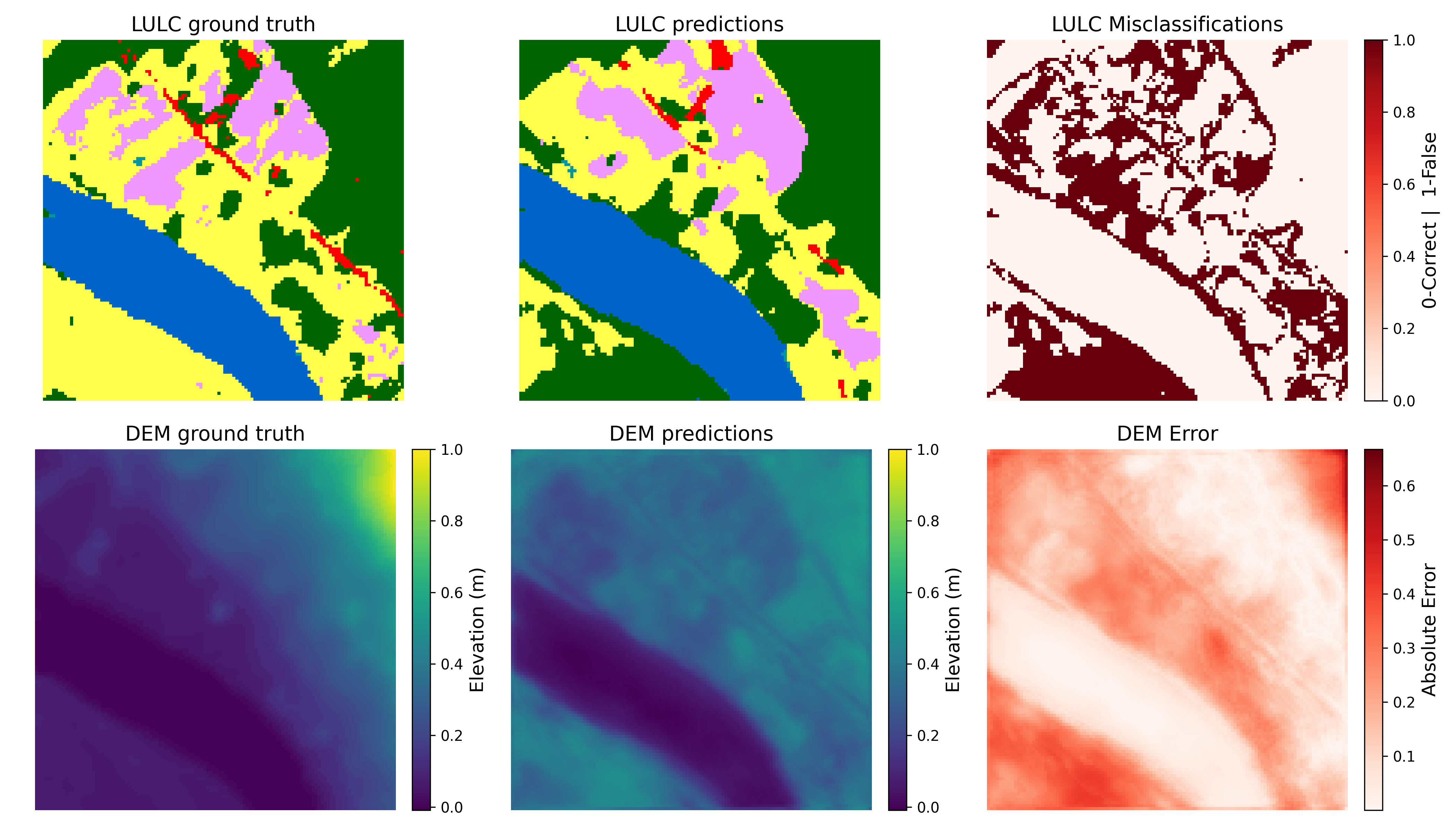}
            \caption{Model predictions and errors, compared against the ground truths, on the \gls{lulc} and \gls{dem} prediction tasks. The predictions are of the best epoch, on two random \benge dataset samples from the test set.}
            \label{fig:lulc_dem_map2}
        \end{figure}

\begin{figure}
            \centering
            \includegraphics[width=0.9\linewidth]{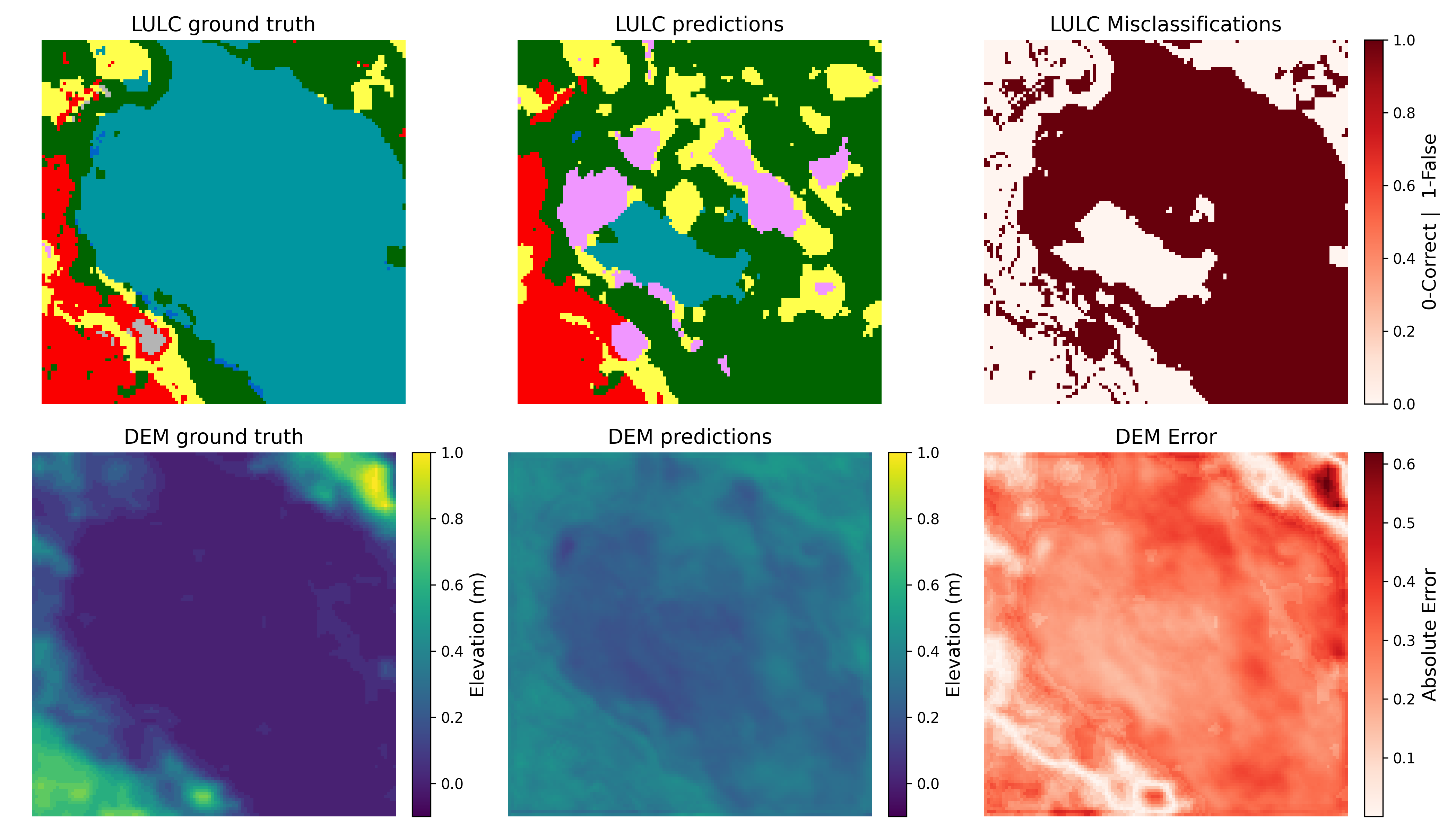}
            \rule{\textwidth}{0.1mm}
            \includegraphics[width=0.9\linewidth]{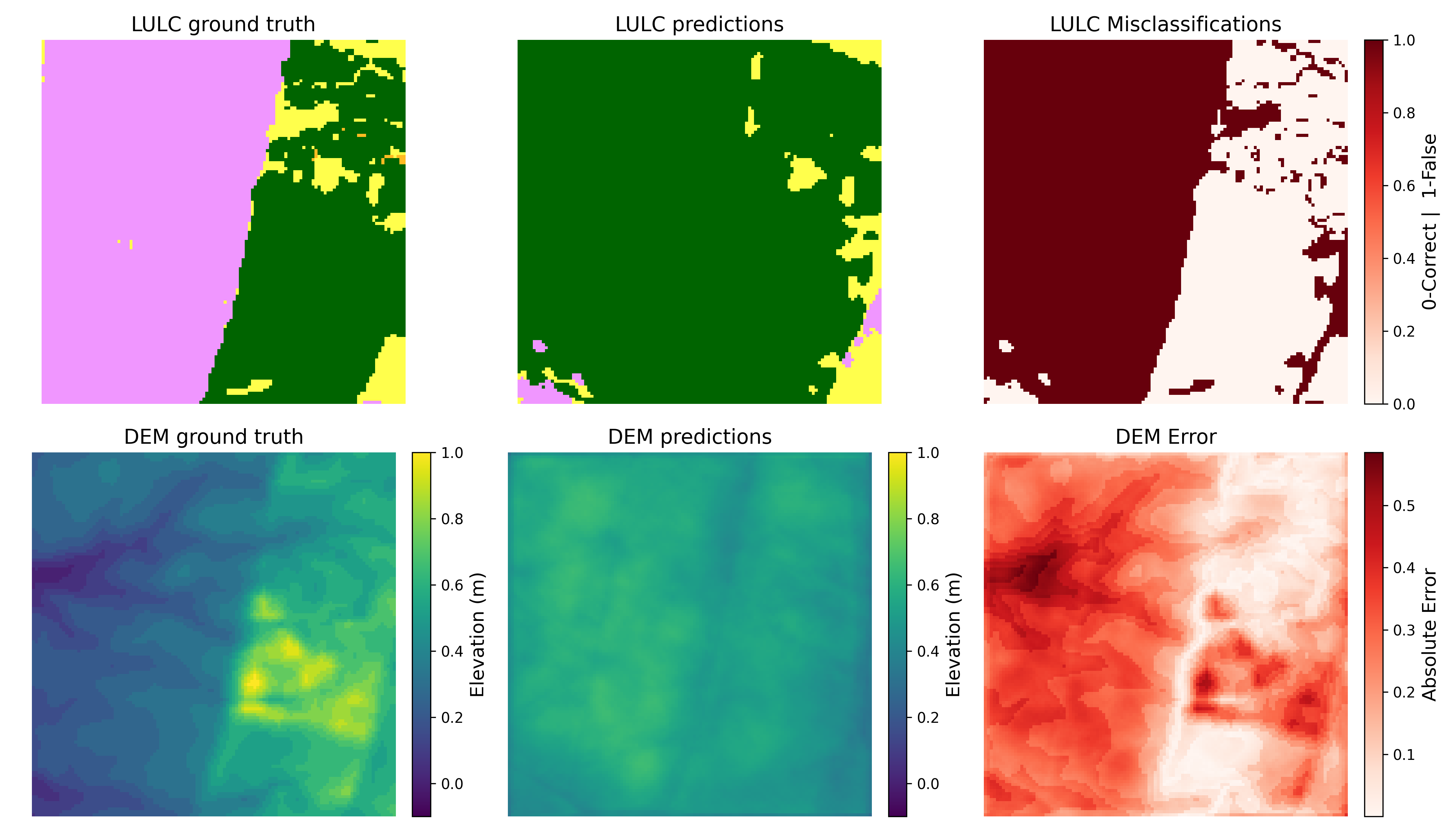}
            \caption{Model predictions and errors, compared against the ground truths, on the \gls{lulc} and \gls{dem} prediction tasks. The predictions are of the best epoch, on two random \benge dataset samples from the test set.}
            \label{fig:lulc_dem_map3}
        \end{figure}

        Figures~\ref{fig:lulc_dem_map2} and \ref{fig:lulc_dem_map3} display ground-truth, predictions and error maps of \gls{lulc} and \gls{dem} tasks. We observe on the four displayed samples how errors from both tasks correlate.  
        
    \subsection{\treesat}\label{app:treesat_xai}
\begin{figure}
            \centering
            \includegraphics[width=\linewidth]{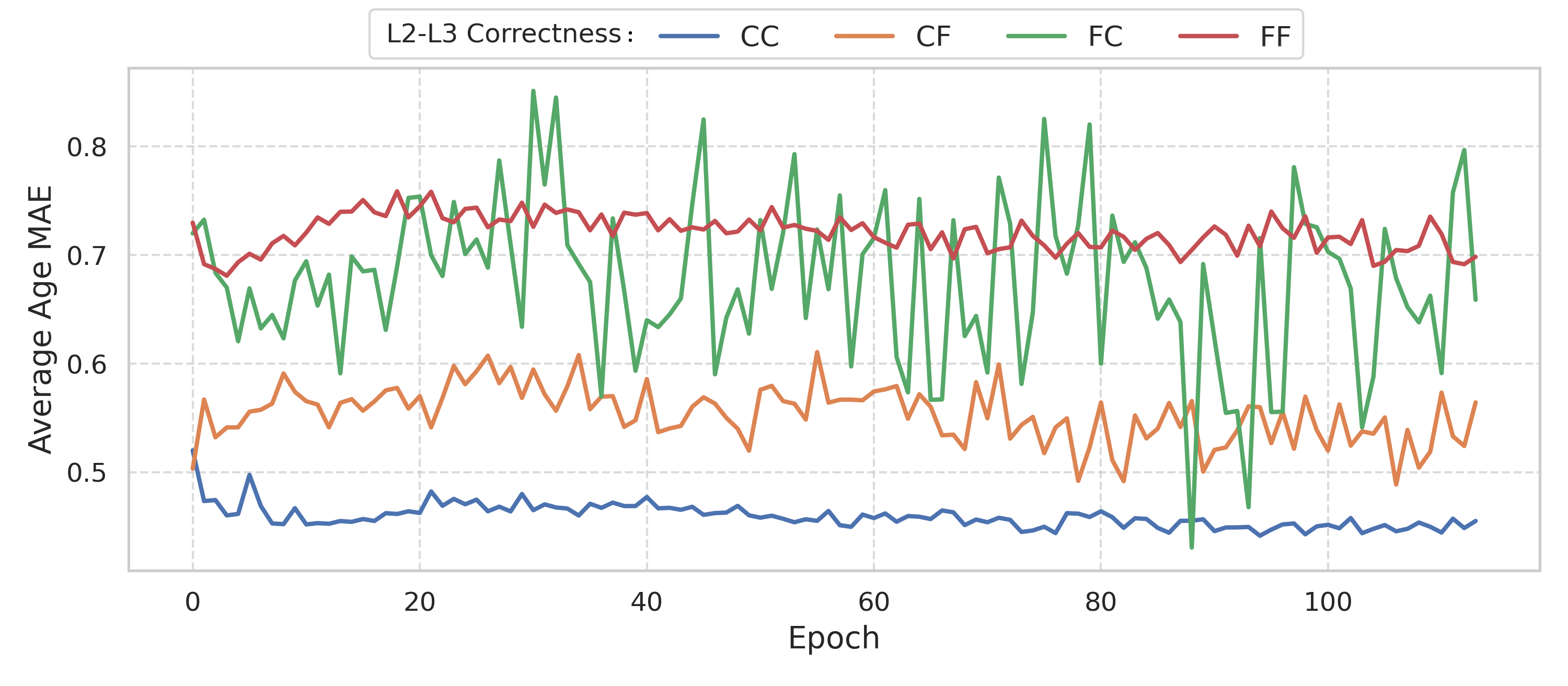}
            \caption{\gls{mae} of the age prediction task, averaged across each combination of correct or false classifications of L2 and L3 labels in the test set of \treesat dataset, throughout the training.}
            \label{fig:treesat_agelineplot}
        \end{figure}

        Experiment \textbf{7}, which we explore in Section~\ref{ssec:xai}, simultaneously predicts L2 label, the tree age, and the primary L3 label. Figure~\ref{fig:treesat_agelineplot} illustrates the average \gls{mae} for age prediction across different combinations of L2 and L3 prediction correctness. 
        The results show that samples with both labels correctly predicted (CC) consistently achieve the lowest \gls{mae} scores throughout the training process. In contrast, samples with both labels incorrectly predicted (FF) consistently exhibit the highest error scores. 
        The intermediate groups, CF and FC, display fluctuating average \gls{mae} values, with CF showing lower error scores compared to FC. This suggests that an incorrect prediction of the L2 label has a negative impact on the accuracy of age prediction, more than an incorrect prediction of the L3 label. Further analysis of the age distribution within each L2 and L3 class may provide additional insights into these observations.
        
\newpage
\bibliographystyle{splncs04}
\bibliography{072-main}

\end{document}